\documentclass[10pt,twocolumn,letterpaper]{article}

\usepackage[pagenumbers]{wacv}      

\usepackage{graphicx}
\usepackage{amsmath}
\usepackage{amssymb}
\usepackage{booktabs}
\usepackage[algo2e]{algorithm2e}
\usepackage{algorithm}
\usepackage{algorithmicx}
\usepackage{algpseudocode}
\usepackage{tcolorbox}
\usepackage{multirow}
\usepackage{makecell}
\usepackage[normalem]{ulem}
\usepackage{amsfonts}
\usepackage{amsthm}
\usepackage{mathrsfs}
\usepackage[title]{appendix}
\usepackage{textcomp}
\usepackage{manyfoot}
\usepackage{listings}
\usepackage{array}
\usepackage{colortbl}

\usepackage[breaklinks,colorlinks]{hyperref}

\usepackage[capitalize]{cleveref}
\crefname{section}{Sec.}{Secs.}
\Crefname{section}{Section}{Sections}
\Crefname{table}{table}{tables}
\crefname{table}{Tab.}{Tabs.}

\begin{document}

\title{Ghost-Connect Net: A Generalization-Enhanced Guidance For Sparse Deep Networks Under Distribution Shifts}

\author{Mary Isabelle Wisell\\
School of Computing and Information Science\\
University of Maine\\
5711 Boardman Hall, \\
Orono, ME 04469, USA\\
{\tt\small mary.wisell@maine.edu}
\and
Salimeh Yasaei Sekeh\\
Department of Computer Science\\
San Diego State University\\
5500 Campanile Drive, \\
San Diego, CA 92182, USA\\
{\tt\small ssekeh@sdsu.edu}
}

\maketitle

\begin{abstract}
   Sparse deep neural networks (DNNs) excel in real-world applications like robotics and computer vision, by reducing computational demands that hinder usability. However, recent studies aim to boost DNN efficiency by trimming redundant neurons or filters based on task relevance, but neglect their adaptability to distribution shifts. We aim to enhance these existing techniques by introducing a companion network, Ghost Connect-Net (GC-Net), to monitor the connections in the original network with distribution generalization advantage. GC-Net's weights represent connectivity measurements between consecutive layers of the original network. After pruning GC-Net, the pruned locations are mapped back to the original network as pruned connections, allowing for the combination of magnitude and connectivity-based pruning methods. Experimental results using common DNN benchmarks, such as CIFAR-10, Fashion MNIST, and Tiny ImageNet show promising results for hybridizing the method, and using GC-Net guidance for later layers of a network and direct pruning on earlier layers. We provide theoretical foundations for GC-Net's approach to improving generalization under distribution shifts.
\end{abstract}

\section{Introduction}
\label{sec1}
Deep neural networks (DNNs) excel across various domains, but face deployment on resource-constrained devices in critical real-world applications since applications like autonomous vehicle navigation \cite{bechtel2018deeppicar, fridman2019advanced, cugurullo2020urban}, simultaneous machine translation \cite{stahlberg2020neural}, and healthcare \cite{yoo2020frequency} demand real-time response without compromising generalization performance \cite{akbari2021does, akbari2022rage}. Sparse DNNs offer a solution by reducing computational demands while maintaining accuracy \cite{wen2016learning, han2020ghostnet, gale2019state, diffenderfer2021winning}. However, they struggle with distribution shift (DS), changes in the input distribution between training and deployment environments. While sparse networks may be efficient, they can lack the adaptability needed to handle DS effectively. This limitation stems from the pruning process, which usually optimizes for performance on a specific dataset or task, potentially removing connections crucial for generalization and limiting their applicability in dynamic environments \cite{gale2019state, gawlikowski2023survey}. 

We investigate how monitoring information flow through network layers can enhance sparse network adaptability to DS, addressing the question:\\
{\it Given tasks with DS, how much does monitoring connectivity between layers pilot the robustness and adaptability of sparse networks?}

We propose Ghost Connect-Net (GC-Net), a novel approach to enhance the generalization capabilities of sparse DNNs under DS. GC-Net acts as a companion network that monitors and guides the connectivity in the original network, offering a more nuanced way to determine which connections to retain. To the best of our knowledge, there are no works which leverage information flow in a partition of the network and utilize the guidance of the downstream connectivity to enhance the performance of the pruned network when facing DS in test data. 

\noindent \textbf{Our Contributions:} 
\begin{enumerate}
    \item We \textbf{introduce GC-Net}, a companion network to guide the original network during pruning.
    \item We \textbf{present the theoretical foundations of GC-Net} and detail its implementation for various architectures, such as ResNet and VGG.
    \item We \textbf{demonstrate GC-Net's effectiveness through experiments} on benchmark datasets, such as CIFAR-10, Fashion MNIST (FMNIST), and Tiny ImageNet (Tiny-IN), as well as variations of these datasets to mimic shifting distributions.
\end{enumerate}

By addressing the DS challenge in sparse networks, this work aims to bridge the gap between sparse model efficiency and generalization across diverse real-world scenarios, potentially expanding their deployment in dynamic environments.

\section{Related Work}\label{sec2}
Recent research has explored various approaches to improve the robustness of sparse networks to DS and domain adaptation. Studies have found that pruned models often struggle more with DS compared to dense counterparts \cite{hooker2019compressed}, leading to the development of pruning techniques that retain connections needed for adapting to new distributions \cite{aditi2022understanding}. Some researchers have focused on developing sparse architectures that dynamically adjust to new data distributions \cite{hendrycks2018benchmarking}, while others have combined sparsity with domain adaptation to preserve features important for domain generalization \cite{koh2021wilds}. A few approaches have been taken to address the DS problem for sparse networks. 

Theoretical insights into the generalization capabilities of sparse networks under DS have been provided \cite{garg2020unified}, and adaptive sparse architectures that reconfigure in response to shifts have been developed \cite{liu2021we}. Test-time training \cite{sun2020test} and fine-tuning with small amounts of target distribution data \cite{kumar2022fine} have been proposed as methods to quickly adapt sparse networks to shifts. Additionally, pruning strategies that preserve connections contributing to invariant features across domains have been introduced \cite{liu2018towards}. The concept of invariant risk minimization has been proposed to learn stable representations across environments \cite{arjovsky2019invariant}, with extensions to out-of-distribution generalization \cite{krueger2021out}. The "lottery ticket hypothesis" has been investigated in the context of DS, revealing that certain sparse sub-networks can maintain robustness across distributions \cite{frankle2020pruning}. A Winning Hand \cite{diffenderfer2021winning} demonstrated that certain compression techniques can inherently improve out-of-distribution robustness. Pruning methods that explicitly consider DS have been developed, aiming to preserve connections that contribute to invariant features \cite{wang2020neural}. 

Research has also explored the impact of layer-wise sparsity on robustness to DS \cite{lee2020layer} and proposed elastic pruning frameworks that allow DNNs to partially recover pruned connections when encountering out-of-distribution data \cite{chen2021elastic}. Building on this work, we propose a novel approach to enhance generalization capabilities of sparse networks for DS by leveraging connectivity to guide network pruning.

\section{Methodology}\label{sec3}
\subsection{Problem Statement}\label{subsec1}
Suppose a deep neural network (DNN) $F^{(L)}$ with $L$ layers is given, which maps the input space $\mathcal{X}$ to a set of classes $\mathcal{T}$, i.e. $F^{(L)}: \mathcal{X}\mapsto \mathcal{T}$. We denote $f^{(l)}$, the $l$-th layer of $F^{(L)}$ with $M_l$ number of filters in layer $l$. The $i$-th filter in layer $l$ is denoted by $f^{(l)}_i$ and $f^{(l)}(x)=\sigma^{(l)}(\omega^{(l)} f^{(l-1)}(x)+b^{(l)})$, where $\sigma ^{(l)}$ is the activation function in layer $l$, $b^{(l)}$ is the offset. In this work, we assume $\sigma$ is bounded. The activation matrix and state of layer $l$ is denoted by $\Delta M_l$ and $A_l$ respectively. 
In this section, we revisit the definition of Pearson correlation ($\rho$), and connectivity matrix ($\mathbf{R}$) between layers. 
The connectivity matrix between two consecutive layers $l$ and $l+1$, with $M_l$ and $M_{l+1}$ filters is defined as
\begin{equation}
\mathbf{R}:=\Big[\rho\left(f^{(l)}_i,f^{(l+1)}_j\right)\Big]_{i=1,\dots,M_l; j=1,\dots,M_{l+1}},
\label{eq:delta}
\end{equation}
where the individual $\rho$ is the connectivity measure between two filters in consecutive layers, so called information flow \cite{andle2022theoretical}. When $\rho$ is Pearson correlation, it can be expressed as:
\begin{equation}
\begin{split}
\rho\left(f^{(l)}_i,f^{(l+1)}_j\right) = & \, cov\left(f^{(l)}_i\;f^{(l+1)}_j\right) \big/\sigma\left(f^{(l)}_i\right)\;\sigma\left(f^{(l+1)}_j\right)
\end{split}
\label{eq:def-connectivity-0}
\end{equation}
where $cov$ is covariance and $\sigma$ denotes standard variation. In classification problem with class variable $Y\in\mathcal{Y}$, input variable $X\in \mathcal{X}$, and joint distribution $(X,Y)\sim D$, when filters are normalized (zero mean and unit variance), the connectivity is simplified by conditional expectation as
\begin{equation}
\begin{split}
\rho\left(f^{(l)}_i,f^{(l+1)}_j\right) 
= & \mathbb{E}_{(X,Y)\sim D}[f^{(l)}(x)f^{(l+1)}(x)|Y].
\end{split}
\label{eq:def-connectivity}
\end{equation}
When $\rho$ is cosine similarity, it is expressed as: 
\begin{equation}
\begin{split}
\rho\left(f^{(l)}_i,f^{(l+1)}_j\right) = & \, \frac{\left(f^{(l)}_i\right)^T \left(f^{(l+1)}_j\right)}{\left\|f^{(l)}_i\right\|_2 \left\|f^{(l+1)}_j\right\|_2},
\end{split}
\label{eq:def-connectivity-cosine}
\end{equation}
where $\|\cdot\|_2$ denotes the L2 norm.
Note that in this paper we consider the absolute value of $\rho$ in the range $[0, 1]$.

\subsection{What is Ghost Connect-Net?}\label{subsec2}
Our goal in this paper is to generate a companion network for the original network $F_O$ that guides the pruning of pre-trained weights in the network, given the connectivity scores of layers introduced in \cref{eq:delta}, so that the predictive power of the sparse network does not degrade performance when shifting the distribution, but also gains a performance improvement. In this section, we first take an in-depth look at the structure of the companion network and show the relationship between the original network and the companion network, so called {\it Ghost-Connect Net (GC-Net)}. Second, we provide a theoretical analysis in which we show that the solution of loss minimization of a sparse network has a solution of engaging the consecutive layers when applying GC-Net guidance, to sparsify the original network. GC-Net is a companion network that stores connectivity-based values of the original network’s layers as weights. These connectivity-weights are pruned, and the pruned locations are mapped back to the original network as pruned connections. GC-Net does no training, relying only on the connectivity-weights calculated from the original network. The current implementation calculates connectivity-weights using Pearson correlation to measure the connectivity between activation states of consecutive layers.

\begin{algorithm}[h!]
    \footnotesize
    \SetAlgoLined 
Set hyperparameters: $\alpha$~(sparsity level), $E$~(epoch), $P$~(pruning method), and $K$~(layer indexes). \\
Given $F_O$, load weights $W_O$ and compute accuracy on ${data}_1$ ($Acc_O$).
\begin{center}
\vspace{-0.3cm}
\begin{tcolorbox}[colback=blue!5!white,colframe=blue!75!black]
\vspace{-.1in}
Create GC-Net:\\
Apply Steps 1-3 shown in Fig.~\cref{fig:GC-Net Steps}\\
         \If{using $F_{GC}Hybrid$}
{Set $K$ layers of GC-Net to not be pruned}
\vspace{-.1in}
\end{tcolorbox}
\end{center}
\begin{center}
\vspace{-0.5cm}
\begin{tcolorbox}[colback=red!5!white,colframe=red!75!black]
\vspace{-.1in}
Prune GC-Net:\\
{\For{layer $l = 1$ \dots, $L$ in $F_{GC}$}
           {\If{$l$ to be pruned}
                {Prune $l$ using $P$-pruning with $sparsity$ level $\alpha$
            }
            {\If{$l$ is Conv or Linear layer and has $weight\_mask$}
                {Set pruned weights to zero and freeze
            }
       }
       }}
\vspace{-.1in}
\end{tcolorbox}
\vspace{-0.4cm}
\end{center}
\begin{center}
\vspace{-0.3cm}
\begin{tcolorbox}[colback=green!5!white,colframe=green!75!black]
\vspace{-.1in}
Map Pruned $F_{GC}$ to $F_O$:\\
        \For{layer $l = 1$ \dots, $L$ in $F_{GC}$}
        {
       {$OrigLayer \gets OrigModules[l]$, 
       $GCLayer \gets GCModules[l]$}\\
            \If{$GCLayer$ has $weight\_mask$}
                {Apply pruning to $origLayer$ using $GCLayer.weight\_mask$\\
                 Freeze pruned weights}
          }
\vspace{-.1in}
\end{tcolorbox}
\vspace{-0.4cm}
\end{center}
\begin{center}
\vspace{-0.3cm}
\begin{tcolorbox}[colback=yellow!5!white,colframe=yellow!75!black]
\vspace{-.1in}
Prune Remaining Layers of $F_O$:\\
       \For{layer $m$ \dots, $L_K$ in $F_O$}
            {Prune $m$-th layer using $P$-pruning with $sparsity$ level $\alpha$\\
            \If{$m$ is Conv2d or Linear and has $weight\_mask$}
        {Set pruned weights to zero and freeze}
      }
\vspace{-.1in}
\end{tcolorbox}
\vspace{-0.5cm}
\end{center}
Fine-tune pruned model for $E$ epochs on ${data}_1$   \\  
Report ${Acc}_1$: pruned model test accuracy on ${data}_1$ \\
Report ${Acc}_2$: pruned model test accuracy on ${data}_2$
    \caption{Pruning DNN with GC-Net Guidance}
    \label{alg:Pruning DNN with GC-Net Guidance}
\end{algorithm}

Throughout the paper,  we denote original (and pre-trained) network, sparse network, and GC-Net by $F_O$, $F_S$, and $F_{GC}$ respectively. The weight matrices of $F_O$ and $F_S$ are $\omega_O$ and $\omega_S$.
GC-Net contains three Steps as follows:

\definecolor{amber}{rgb}{1.0, 0.75, 0.0}
\definecolor{applegreen}{rgb}{0.55, 0.71, 0.0}
\definecolor{azure}{rgb}{0.0, 0.5, 1.0}
\definecolor{babyblue}{rgb}{0.54, 0.81, 0.94}
\begin{figure*}[t]
    \centering
    \includegraphics[width=\textwidth]{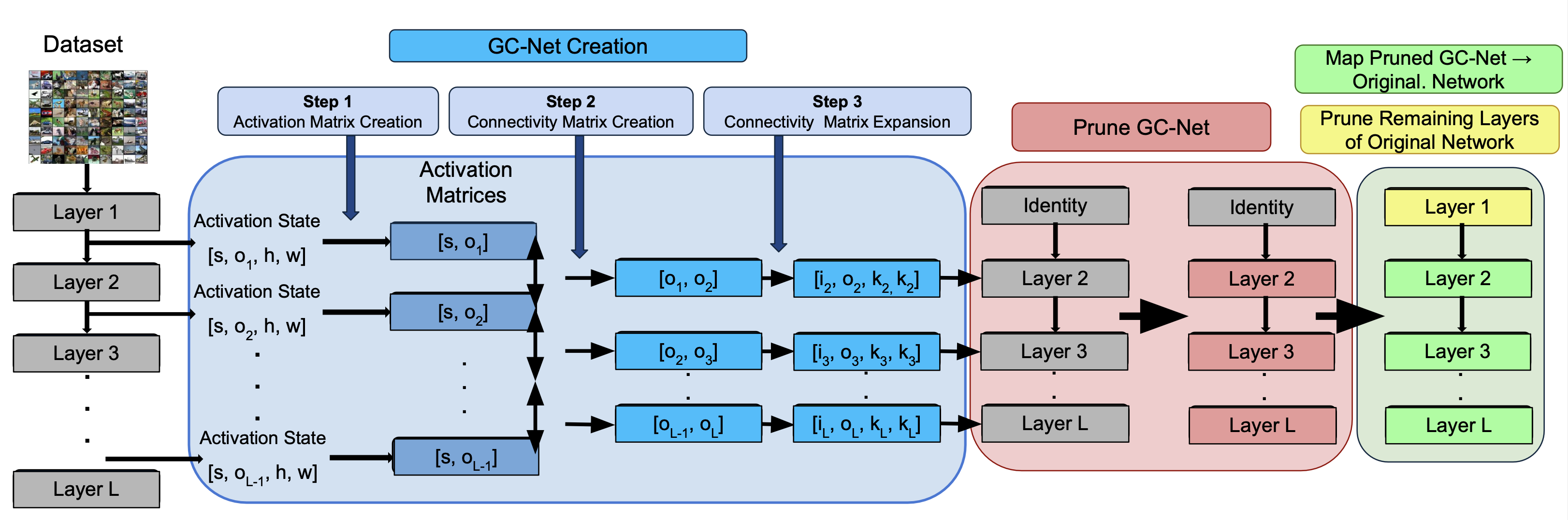}
    \caption{GC-Net Guidance Overview: \textcolor{azure}{(Blue module)} Generate activation matrices (step 1), create connectivity matrices via Pearson Correlation (step 2), expand matrices as GC-Net weights (step 3). \textcolor{red}{(Red module)} Prune GC-Net. \textcolor{applegreen}{(Green module)} Map pruned GC-Net to original network. \textcolor{amber}{(Yellow module)} Prune remaining original network layers.}
    \label{fig:GC-Net Steps}
\end{figure*}
\uline{\textit{ Step 1 - Activation Matrix Calculation:}} Each sample in the training dataset, $S$, is passed through the trained network, and the activation state is collected for each layer. For each layer $l$ in the $L$-layer network, a four-dimensional matrix storing the activation state of each sample $[s, o, h, w]$ is initially formed, where $s$ represents samples, $o$ output channels, and $h, w$ the height and width of the activation state $A$. This is then reduced to a two-dimensional activation matrix $[s, o]$ by averaging over $h$ and $w$.

\uline{\textit{Step 2 - Connectivity Matrix Calculation:}} To calculate the connectivity matrices $\mathbf{R}$ from the activation matrices $\Delta M$, the $\rho$ value in ~\cref{eq:def-connectivity} is calculated for the columns of consecutive $\Delta M$ matrices. The column-wise Pearson correlation applied to the $\Delta M$ matrices produces connectivity matrices with dimensions $[o_{l+1}, o_l]$, as illustrated in step 2 of \cref{fig:GC-Net Steps}. Since the $\mathbf{R}$ matrices are calculated between pairs of $\Delta M$ matrices, the number of $\mathbf{R}$ matrices is one less than the number of $\Delta M$ matrices, producing $L-1$ connectivity matrices ($L:$ \# of layers).

\uline{\textit{Step 3 - Reshaping Connectivity Matrices:}} To load the connectivity matrices as weights in GC-Net, they are transposed and expanded to be the same dimensions as layer $l+1$ in the original network. For example, assume a connectivity matrix exists for the connections between layers $l$ and $l+1$ in the original network, where layer $l$ has dimensions $[a, b, c, c]$ and layer $l+1$ has dimensions $[a, g, c, c]$. The connectivity matrix for these layers will initially have dimensions $[g, a]$, and will be transposed to be $[a, g]$, then expanded to have dimensions $[a, g, c, c]$. 

The pseudo code for steps 1-3 (GC-Creation) is provided in Appendix A \cref{subsec9}. 

Once the $\mathbf{R}$ matrices have been created and expanded, they can be loaded as the {\em weights} of $F_{GC}$ denoted by $W_{GC}$. An untrained network with an identical architecture to the original network is created, and the first layer in the network is changed to an identity layer (as there is $L-1$ connectivity matrices). 

\subsubsection{GC-Net for ResNet}\label{subsubsec1}
\begin{figure}[ht]
  \centering
  \includegraphics[width=0.95\linewidth]{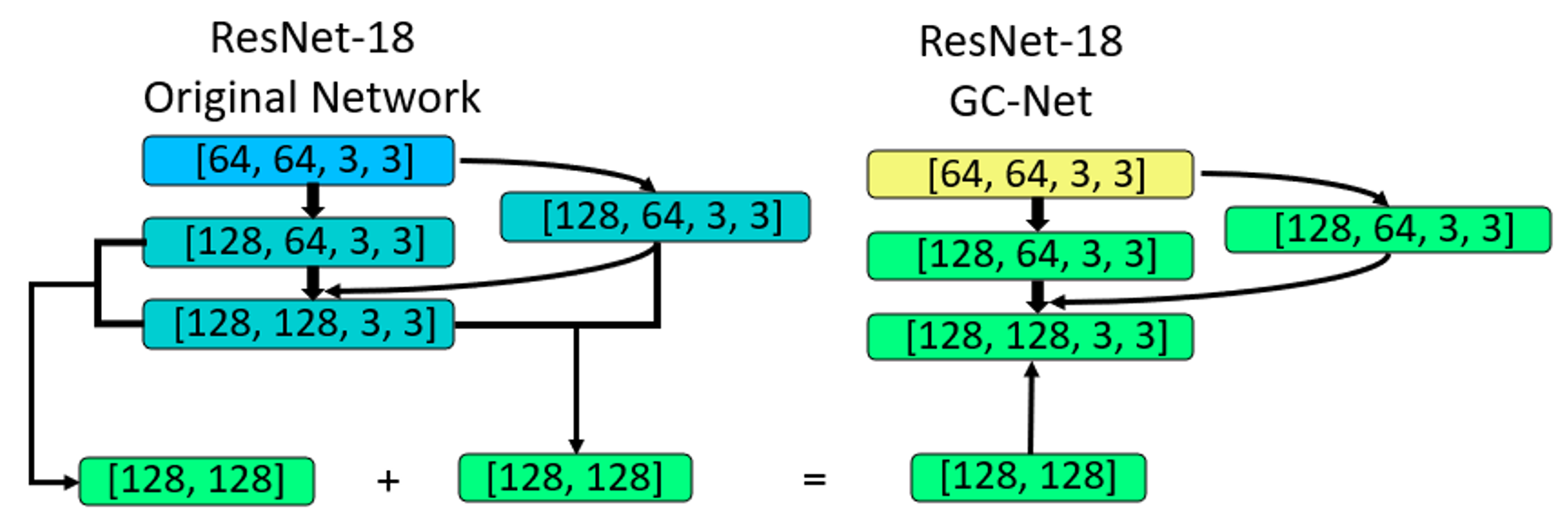}
  \caption{\small When the outputs of two layers converge into a single subsequent layer, their respective connectivity matrices are added. This combined connectivity matrix is then treated as a single matrix and loaded as weights for the target layer in GC-Net.}
  \label{fig:ResNet Exception}
\end{figure}

For architectures with skip layers, the output of two layers may combine to feed into a single layer. This results in two connectivity matrices assigned to a single layer in $F_{GC}$. As shown in \cref{fig:ResNet Exception}, GC-Net addresses this case by adding the $\mathbf{R}$ matrices together, and the resulting sum is then processed as a single connectivity matrix. is mimics the behavior of the original skip layer where the output of the skip layer is added to the next layer's input.

\subsubsection{GC-Net for VGG}\label{subsubsec2}
In VGG architectures, there is an average pooling layer between the final convolutional layer and the first linear layer, where the number of input channels in layer $l+1$ differs from the number of output channels of layer $l$. Here, the the connectivity matrix is transposed to have dimensions $[o_l, o_{l+1}]$, and expanded only along the second dimension, $o_{l+1}$, where the values in each column are duplicated $c_p$ times, where $c_p$ is the size of the kernel for the pooling layer, $p$. The expanded values are then moved to the locations where they would have been pooled together by the kernel for the pooling layer.

\subsubsection{GC-Net for Hybrid}\label{subsubsec3}
\begin{figure}[h]
\centering
\includegraphics[width=0.9\linewidth]{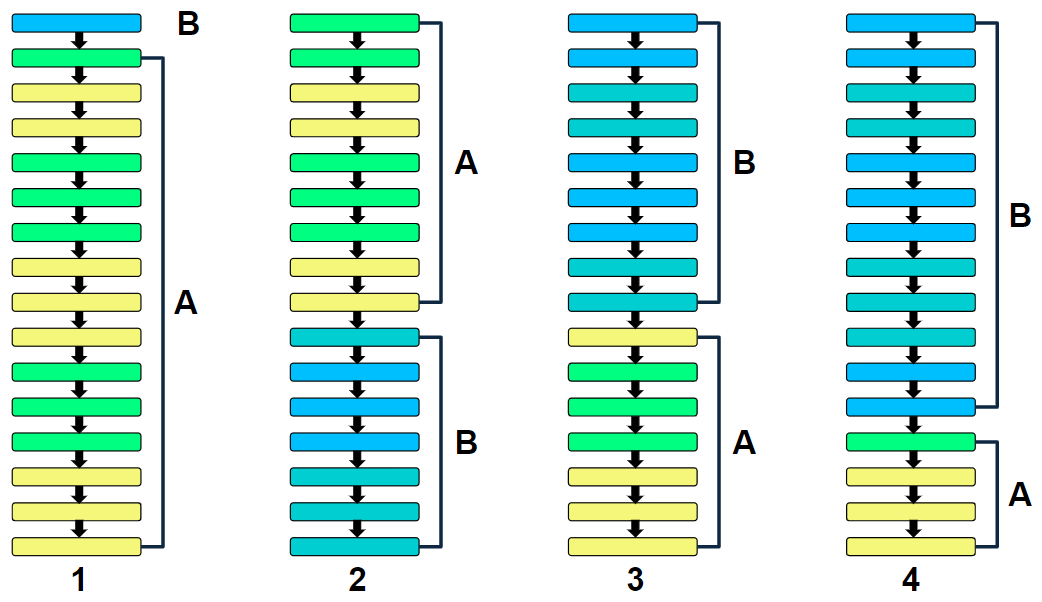}
\caption{VGG16-BN Hybrid: Part B is pruned directly in original network $F_O$ and Part A is pruned by guiding GC-Net $F_{GC}$. Models (1-4, from left) show: 
\newline 1) \textbf{Full GC-Net}: the first layer is pruned directly and all other layers are pruned with GC-Net
\newline 2) \textbf{CG-Net - FH}: the last 50\% is pruned directly and the first 50\% is pruned with GC-Net
\newline 3) \textbf{GC-Net - BH}: the first 50\% is pruned directly and the last 50\% is pruned with GC-Net
\newline 4) \textbf{GC-Net-B25\%}: the first 75\% is pruned directly and the last 25\% is pruned with GC-Net}
\label{fig:Hybrid Examples}
\end{figure}

We define GC-Net hybrid architectures as networks that apply the GC-Net method to one part of the original network $F_O$, and then apply the selected pruning method directly to the remaining part. Examples of current GC-Net hybrid architectures are demonstrated in \cref{fig:Hybrid Examples}: Full GC-Net, GC-Net - Front Half (GC-Net - FH), GC-Net - Back Half (GC-Net - BH), and GC-Net - Back25\% (GC-Net - B25\%). We denote GC-Net hybrid by $F_{GC}Hybrid$.

\section{Experimental Study}\label{sec4}
\subsection{Baselines, Datasets, and Setup}\label{subsec3}
We use two baseline architectures: ResNet-18 \cite{he2016deep} and VGG16-BN \cite{simonyan2014very}. These architectures are pre-trained on three datasets: CIFAR-10, FMNIST resized to 32x32, and Tiny-IN. Baseline models were established by fine-tuning pretrained models using stochastic gradient descent (SGD) with varying learning rates and epochs: {\it ResNet-18:} CIFAR-10 (from \cite{phan2021cifar10}, 10 epochs), FMNIST (PyTorch, 15 epochs), Tiny-IN (PyTorch, 5 epochs), {\it VGG16-BN:} All from PyTorch; CIFAR-10 (35 epochs), FMNIST (10 epochs), Tiny-IN (5 epochs). 

Learning rates ranged from 0.01 to 0.0001 and were adjusted during training for each model. Additional results for ResNet-18 and VGG16-BN for L1-norm, L2-norm, C-SNIP, and OS-SynFlow for additional hybrid methods are provided in Appendix A in \cref{subsec7} for Pearson Correlation results and \cref{subsec8} for cosine similarity.

To test robustness to DS, we evaluate each model on the proposed modified versions of the baseline datasets. The modified datasets include:

\begin{itemize}     
    \item \textbf{CJG (Color Jittering and Geometry)}: Introduces color jittering, random rotations, and affine transformations.
    \item \textbf{RNB (Random Noise and Blur)}: Simulates sensor noise and focus issues through Gaussian noise and blur.
    \item \textbf{LO (Lighting and Occlusion)}: Creates lighting variations through color jittering and occlusions with random patches.
\end{itemize}

We perform various pruning trials for all four GC-Net hybrid models described in \cref{fig:Hybrid Examples}, and the original pretrained model as a baseline. For the hybrid model methods, the GC-Net portion of the model is pruned first, and the remaining $K$ layers of the model are pruned directly in the original network $F_O$.

The model is pruned to a fixed sparsity level $\alpha$ (20\%, 40\%, 60\%, or 80\%) using one of four pruning methods $P$:

\begin{itemize}
    \item \textbf{L1-norm}: The weights are pruned based on their L1-norm, and each layer is pruned to sparsity level $\alpha$.
    \item \textbf{L2-norm}: The weights are pruned based on their L2-norm, and each layer is pruned to sparsity level $\alpha$.
    \item \textbf{OS-SynFlow (Oneshot SynFlow)}: A variation of the SynFlow \cite{tanaka2020pruning} pruning method, modified to be a one-shot pruning method, where each layer is pruned to sparsity level $\alpha$ based on the SynFlow scores.
    \item \textbf{C-SNIP (Capped SNIP)}: A variation of the SNIP \cite{lee2018snip} pruning method, where the method is limited to not prune more than 95\% of any single layer to achieve the target $\alpha$ while avoiding layer collapse.
\end{itemize}
Once the model is pruned, it is fine tuned for $E = 10$ epochs on dataset ${data}_1:=\{(x_i,y_i\}_{i=1}^{N_1}$ from random vector $(\mathbf{X},Y)\sim D_1$ with distribution $D_1$ and label set $\mathcal{Y}_1$ using SGD and a learning rate of 0.0001, and the final test accuracy is recorded. The model is then tested on a shifted distribution dataset, dataset ${data}_2:=\{(x_i,y_i\}_{i=1}^{N_2}$ from random vector $(\mathbf{X},Y)$ with distribution $D_2$ and label set $\mathcal{Y}_2$, and the test accuracy is recorded. Experiments are repeated for 3 trials. 

\subsection{Domain and Distribution Shift Results:}\label{subsec4}
We present a detailed analysis of the performance of GC-Net-guided pruning vs traditional pruning techniques on various architectures. Our results highlight the effectiveness of the GC-Net methodology across different datasets and distribution shifts. The results in each table represent the average final accuracy after 10 fine-tuning epochs across 3 trials. The full results for all pruning methods for ResNet-18 and VGG16-BN across all sparsities can be found in Appendix A \cref{secA1} and indicate all hybrid methods appear to maintain competitive accuracy at higher sparsity levels. \cref{tab:performance_comparison_resnet18_pruning_methods} and \cref{tab:performance_comparison_vgg16bn_pruning_methods} demonstrate the top performing hybrid methods, GC-Net - BH and GC-Net - B25\%, for all pruning methods at 20\% sparsity for the CIFAR-10 dataset and variations. 

For ResNet-18, \cref{tab:performance_comparison_resnet18_pruning_methods} indicates GC-Net-BH and GC-Net-B25\% exhibit strong performance in comparison to the traditional pruning methods. GC-Net - B25\% consistently performs well across all pruning techniques, often outperforming the original model in distribution shift datasets, such as CJG shift. GC-Net - BH demonstrates similarly competitive performance. While it trails behind GC-Net - B25\%, it still consistently outperforms the original pruning method on its own for the C-SNIP and OS-SynFlow experiments. Both the GC-Net - BH and GC-Net - B25\% methods outperformed the original method for almost all pruning methods for the LO distribution shift. For VGG16-BN, \cref{tab:performance_comparison_vgg16bn_pruning_methods} GC-Net-B25\% frequently outperforms the original pruning method on its own for the original and RNB datasets. For VGG16-BN, GC-Net - BH offers equally competitive results,  and is the best performing model for the C-SNIP CJG and RNB variations. While the results for \cref{tab:performance_comparison_resnet18_pruning_methods,tab:performance_comparison_vgg16bn_pruning_methods} may appear slightly inconsistent, it is important to consider the influence of the various hyper-parameters in the experiments. The accuracy is highly dependent on the sample size, dataset, and distribution shift variation. Thus, the optimal sparsity level is often inconsistent between datasets and pruning methods.

Overall, the hybrid models exhibit different behavior depending on the type of domain shift. For the CJG and RNB distribution shifts, the hybrid models show a strong performance boost over the original model. For the Lo distribution shift, all methods, including the original pruning method alone, show a significant drop in performance. However, the hybrid method outperform the original method alone for the LO shift in most cases. Additionally, the overall performance of all hybrid methods, including the original pruning method alone, was lower for the L1-Norm and L2-Norm pruning methods than for C-SNIP or OS-SynFlow. However, the hybrid models more frequently outperform the original method alone, for the C-SNIP and OS-SynFlow methods, indicating the hybrid methods offer a boost in performance for the higher performing pruning methods for both VGG16-BN and ResNet-18 architectures. Additionally, the variations between the performance of the hybrid methods between the VGG16-BN and ResNet-18 results indicate there may be opportunities for further investigation into the optimal hybrid model for pruning each architecture or for each dataset.

\begin{table}[!ht]
\centering
\footnotesize
\setlength{\tabcolsep}{4pt}
\begin{tabular}{|>{\centering\arraybackslash}p{1.8cm}|c|cccc|}
\hline
\multirow{2}{*}{\makecell[c]{Pruning\\Method}} & \multirow{2}{*}{Hybrid Type} & \multicolumn{4}{c|}{CIFAR-10 DS Dataset} \\
 &  & Orig($Acc_1$) & CJG & RNB & LO \\
\hline
\multirow{3}{*}{\makecell[c]{L1-Norm\\$Acc_O$: 93.14\%}} 
& GC-Net-BH & 93.05 & 73.52 & 23.63 & \textbf{13.82} \\
& GC-Net-B25\% & \textbf{93.11} & 73.32 & 24.24 & 13.75 \\
& Original & \cellcolor{yellow!25}93.04 & \cellcolor{yellow!25}\textbf{74.38} & \cellcolor{yellow!25}\textbf{24.49} & \cellcolor{yellow!25}13.69 \\
\hline
\multirow{3}{*}{\makecell[c]{L2-Norm\\$Acc_O$: 93.14\%}} 
& GC-Net-BH & 91.76 & 70.89 & 23.23 & 13.74 \\
& GC-Net-B25\% & 91.85 & 70.63 & 23.46 & \textbf{14.15} \\
& Original & \cellcolor{yellow!25}\textbf{91.99} & \cellcolor{yellow!25}\textbf{71.62} & \cellcolor{yellow!25}\textbf{24.3} & \cellcolor{yellow!25}13.75 \\
\hline
\multirow{3}{*}{\makecell[c]{C-SNIP\\$Acc_O$: 93.14\%}} 
& GC-Net-BH & \textbf{93.14} & \textbf{73.82} & \textbf{24.76} & \textbf{13.74} \\
& GC-Net-B25\% & \textbf{93.14} & 73.75 & 24.51 & \textbf{13.74} \\
& Original & \cellcolor{yellow!25}93.08 & \cellcolor{yellow!25}73.77 & \cellcolor{yellow!25}24.62 & \cellcolor{yellow!25}13.64 \\
\hline
\multirow{3}{*}{\makecell[c]{OS-SynFlow\\(Pearson)\\$Acc_O$: 93.14\%}} 
& GC-Net-BH & 93.10 & 73.41 & 24.43 & 13.65 \\
& GC-Net-B25\% & 93.09 & \textbf{74.16} & \textbf{24.67} & \textbf{13.92} \\
& Original & \cellcolor{yellow!25}\textbf{93.15} & \cellcolor{yellow!25}73.87 & \cellcolor{yellow!25}24.48 & \cellcolor{yellow!25}13.32 \\
\hline
\multirow{3}{*}{\makecell[c]{OS-SynFlow\\(Cosine)\\$Acc_O$: 93.14\%}} 
& GC-Net-BH & 93.01 & 73.93 & 25.65 & 13.82 \\
& GC-Net-B25\% & 93.02 & \textbf{73.97} & \textbf{26.26} & \textbf{14.01} \\
& Original & \cellcolor{yellow!25}\textbf{93.15} & \cellcolor{yellow!25}73.87 & \cellcolor{yellow!25}24.48 & \cellcolor{yellow!25}13.32 \\
\hline
\end{tabular}
\caption{ResNet-18 average accuracy (\%) across three trials for all pruning methods for CIFAR-10 at 20\% sparsity level. The results of the original pruning method alone, without GC-Net are highlighted in yellow.}
\label{tab:performance_comparison_resnet18_pruning_methods}
\end{table}

\begin{table}[!ht]
\centering
\setlength{\tabcolsep}{4pt}
\footnotesize
\begin{tabular}{|>{\centering\arraybackslash}p{1.8cm}|c|cccc|}
\hline
\multirow{2}{*}{\makecell[c]{Pruning\\Method}} & \multirow{2}{*}{Hybrid Type} & \multicolumn{4}{c|}{CIFAR-10 DS Dataset} \\
 &  & Orig($Acc_1$) & CJG & RNB & LO \\
\hline
\multirow{3}{*}{\makecell[c]{L1-Norm\\$Acc_O$: 93.18\%}} 
& GC-Net-BH & 93.2 & \textbf{76.53} & 28.93 & 13.13 \\
& GC-Net-B25\% & 93.13 & 76.45 & 29.72 & 13.32 \\
& Original & \cellcolor{yellow!25}\textbf{93.23} & \cellcolor{yellow!25}76.31 & \cellcolor{yellow!25}\textbf{29.8} & \cellcolor{yellow!25}\textbf{13.34} \\
\hline
\multirow{3}{*}{\makecell[c]{L2-Norm\\$Acc_O$: 93.18\%}} 
& GC-Net-BH & 92.1 & 74.09 & \textbf{29.51} & \textbf{12.8} \\
& GC-Net-B25\% & \textbf{92.32} & \textbf{74.41} & 29.14 & 12.78 \\
& Original & \cellcolor{yellow!25}92.24 & \cellcolor{yellow!25}73.92 & \cellcolor{yellow!25}28.6 & \cellcolor{yellow!25}12.67 \\
\hline
\multirow{3}{*}{\makecell[c]{C-SNIP\\(Pearson)\\$Acc_O$: 93.18\%}} 
& GC-Net-BH & 93.08 & 76.44 & \textbf{30.16} & 13.39 \\
& GC-Net-B25\% & \textbf{93.32} & \textbf{76.61} & 29.95 & 12.86 \\
& Original & \cellcolor{yellow!25}93.23 & \cellcolor{yellow!25}76.41 & \cellcolor{yellow!25}29.62 & \cellcolor{yellow!25}\textbf{13.18} \\
\hline
\multirow{3}{*}{\makecell[c]{C-SNIP\\(Cosine)\\$Acc_O$: 93.18\%}} 
& GC-Net-BH & 92.95 & \textbf{76.55} & \textbf{29.83} & 12.91 \\
& GC-Net-B25\% & \textbf{93.29} & 76.29 & 29.77 & \textbf{13.72} \\
& Original & \cellcolor{yellow!25}93.23 & \cellcolor{yellow!25}76.41 & \cellcolor{yellow!25}29.62 & \cellcolor{yellow!25}13.18 \\
\hline
\multirow{3}{*}{\makecell[c]{OS-SynFlow\\$Acc_O$: 93.18\%}} 
& GC-Net-BH & 93.06 & 76.54 & 29.8 & 12.88 \\
& GC-Net-B25\% & \textbf{93.28} & 76.5 & \textbf{30.15} & 12.81 \\
& Original & \cellcolor{yellow!25}93.23 & \cellcolor{yellow!25}\textbf{76.66} & \cellcolor{yellow!25}29.59 & \cellcolor{yellow!25}\textbf{13.65} \\
\hline
\end{tabular}
\caption{VGG16-BN average accuracy (\%) across three trials for all pruning methods for CIFAR-10 at 20\% sparsity. The results of the original pruning method alone, without GC-Net are highlighted in yellow.}
\label{tab:performance_comparison_vgg16bn_pruning_methods}
\end{table}

\subsection{FLOPs Analysis}\label{subsec5}

\begin{figure*}[htb!]
 \centering 
 \begin{subfigure}{0.32\linewidth}
  \centering
  \includegraphics[width=\linewidth,height=0.25\textheight,keepaspectratio]{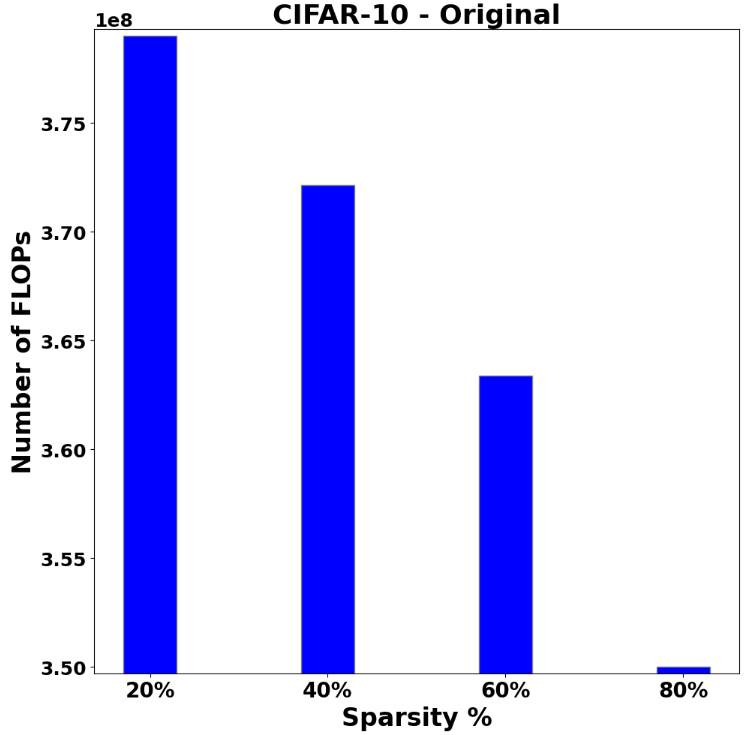}
  \caption{\footnotesize FLOPs when pruning with the original method and no use of GC-Net.}
  \label{fig:ResNet18 Original OS-SynFlow FLOPs}
 \end{subfigure}
 \hfill
 \begin{subfigure}{0.32\linewidth}
  \centering
  \includegraphics[width=\linewidth,height=0.25\textheight,keepaspectratio]{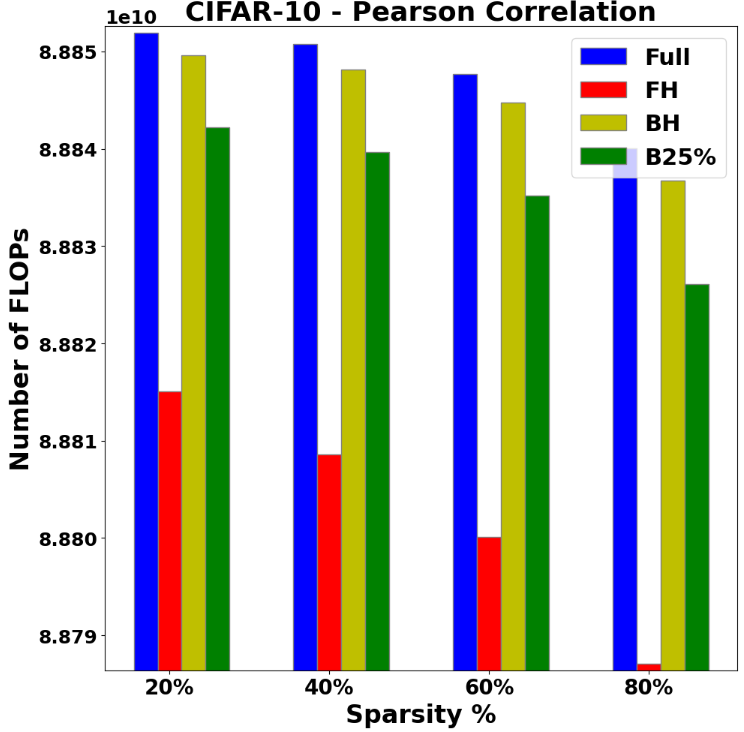}
  \caption{\footnotesize FLOPs when pruning with Pearson correlation for GC-Net.}
  \label{fig:ResNet18 Pearson OS-SynFlow FLOPs}
 \end{subfigure}
 \hfill
 \begin{subfigure}{0.32\linewidth}
  \centering
  \includegraphics[width=\linewidth,height=0.25\textheight,keepaspectratio]{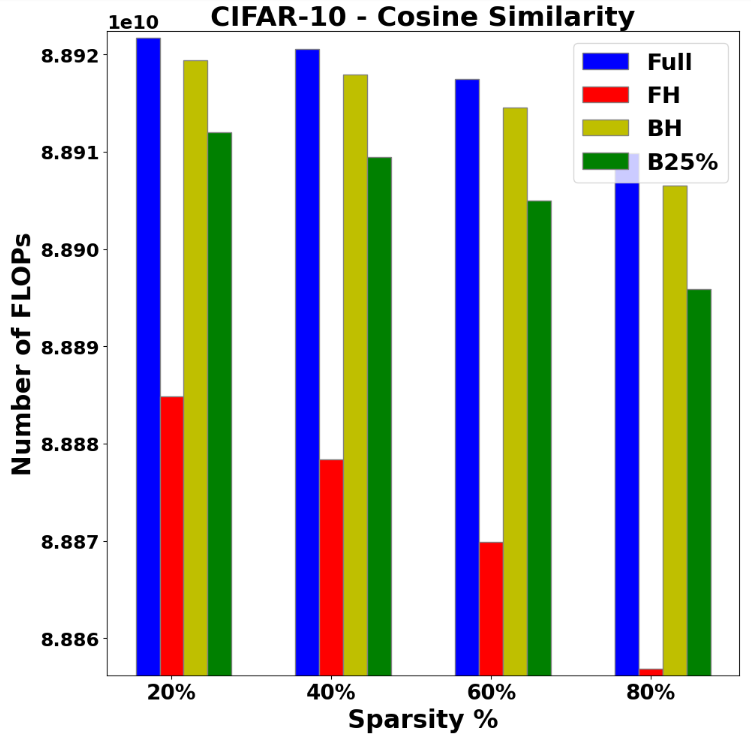}
  \caption{\footnotesize FLOPs when pruning with cosine similarity for GC-Net.}
  \label{fig:ResNet18 Cosine OS-SynFlow FLOPs}
 \end{subfigure}
 
 \caption{\footnotesize Comparison of ResNet-18 OS-SynFlow FLOPs for original and hybrid methods using different similarity metrics for CIFAR-10.}
 \label{fig:ResNet18 OS-SynFlow FLOPs}
\end{figure*}

The FLOPs required for GC-Net can be broken down into three main components: 

\begin{itemize}
    \item \textbf{Connectivity Matrix Calculation:} Computing connectivity between layer activations (most FLOPs-intensive)
    \item \textbf{GC-Net Pruning:} Less intensive than connectivity calculations
    \item \textbf{Mapping Pruned Connections:} Transferring pruning decisions to the original network (least intensive)
\end{itemize}
While GC-Net requires substantial FLOPs, it's only created once at the beginning of the pruning process and not used during inference. The improved pruning decisions can lead to sparser networks that maintain performance under distribution shifts. As shown in \cref{fig:ResNet18 Pearson OS-SynFlow FLOPs,fig:ResNet18 Cosine OS-SynFlow FLOPs}, GC-Net uses significantly more FLOPs than the original pruning method alone \cref{fig:ResNet18 Original OS-SynFlow FLOPs}. However, most FLOPs are used for initial connectivity matrix calculations. For iterative pruning, or methods requiring multiple candidate evaluations, like EagleEye \cite{li2020eagleeye}, the cost of creating GC-Net becomes negligible compared to repeated pruning technique applications, while offering increased DS performance.

\subsection{Pearson Correlation vs Cosine Similarity}\label{subsec6}
The choice of connectivity measurement in GC-Net significantly impacts performance and computational cost. Our experiments reveal trade-offs between cosine similarity and Pearson correlation. Cosine similarity demonstrated a slight increase in performance in certain scenarios. As shown in \cref{tab:performance_comparison_resnet18_pruning_methods} the performance of the hybrid methods with OS-SynFlow using cosine similarity had slightly better overall performance than those with Pearson correlation. However, this trend is less apparent with VGG16-BN and C-SNIP in \cref{tab:performance_comparison_vgg16bn_pruning_methods}. This indicates that the optimal connectivity metric may vary with architecture or dataset. There is opportunity for additional investigation into the optimal connectivity metric for each architecture, dataset, or pruning technique. Further research into alternative connectivity measurements, such as chordal distance or subspace collinearity \cite{yi2011user}, could yield insights into performance and computational cost trade-offs. The additional results for cosine similarity for OS-SynFlow for ResNet-18 and C-SNIP for VGG16-BN can be found in Appendix A \cref{subsec8}.

\section{Theoretical Explanation}\label{sec5}
A basic DNN with $L$ layers is described by $F^{(L)} = f^{(L)}(f^{(L-1)}(\dots f^{(1)}(x^{(0)}))\dots )$, and each individual layer can be defined $f^{(l)}(x^{(l-1)}) = \sigma^{(l)}(w^{(l)}x^{(l-1)} + b^{(l)})$ for $l=1,\ldots, L$, where $x^{(l-1)}$ defines the input activation to layer $l$. In addition, we define a smaller subset of the network between layers $i~\text{and}~j$, where $j > i$ as $ G^{(i,j)} = f^{(j)}(f^{(j-1)}(\dots f^{(i)}(x^{(i-1)})\dots )$. Define $s^{(L)}$ as the layer importance scores for layer $L$, the last layer, we solve the following optimization problem:
\vspace{-1ex}
\begin{gather}
\begin{aligned}\label{eq1:objective}
\mathop{argmin}\limits_{\tilde{s}^{(l)}} & \sum_{n = 1}^{N} \mathcal{F}(\tilde{s}^{(l)} \mid x^{(l)}_n, s^{(L)}; G^{(l+1,L)}), 
\end{aligned}
\end{gather}

\vspace{-1ex}
where $\mathcal{F}$ is defined as:

\begin{gather}
\begin{aligned}\label{eq2:objective}
\varphi(s^{(L)}, |\rho(G^{(l+1,L)}(x^{(l)}_n)) - \rho(G^{(l+1,L)}(\tilde{s}^{(l)}\odot x^{(l)}_n))| 
\end{aligned}
\end{gather}

where $\odot$ is element-wise product, $|.|$ is element-wise absolute and $\|.\|$ is $L_2$-norm. In \cref{eq2:objective} $\varphi$ is a {\it propagate} function and $\rho$ is {\it connectivity} passing through sub-network. In GC-Net we use the particular example of $\varphi$ and $\rho$, dot product and uniform function, respectively:
\begin{multline}
\mathcal{F}(\tilde{s}^{(l)} \mid x^{(l)}_n, s^{(L)}; G^{(l+1,L)}) = \\
\langle s^{(L)}, \vert G^{(l+1,L)}(x^{(l)}_n) - G^{(l+1,L)}(\tilde{s}^{(l)}\odot x^{(l)}_n) \vert \rangle
\label{eq:objective_breakdown}
\end{multline}
where $\langle. ,.\rangle$ is dot product. To solve the optimization problem \cref{eq1:objective}, we use the information flow, $r^{i\rightarrow j}$, from layer $f^{(i)}$ to layer $f^{(j)}$. Define the operation $\Delta$ as an operation that integrated all propagate information (connectivity) from previous layers, i.e. $r^{i\rightarrow j}\Delta\; r^{j\rightarrow k}$ means that information from sub-network $G^{(i,j)}$ is integrated in information from sub-network $G^{(j,k)}$. Therefore the information flow between layers in $G^{(i,j)}$ as follows

\begin{align}
r^{i \rightarrow j} = \rho(G^{(i,j)}(x)) = \rho(G^{(i,j-1)}(x)) \; \Delta \; r^{(j-1) \rightarrow j}(x).
\end{align}

To clarify the operation $\Delta$ in GC-Net, suppose we have three consecutive layers $f^{(l)}$, $f^{(l+1)}$ and $f^{(l+2)}$ with $M_l$, $M_{l+1}$ and $M_{l+2}$ output channels. Then $r^{l\rightarrow l+1}$ is a matrix of size $M_l\times M_{l+1}$ and $r^{l+1\rightarrow l+2}$ is a matrix of size $M_{l+1}\times M_{l+2}$. We set $\Delta$ operation in $F_{GC}$ such that the information flow $r^{l\rightarrow l+2}$ is a matrix of size $M_{l+(l+1)+(l+2)}\times M_{l+1}$. 
Going back to ~\cref{eq2:objective}, for two entries $x$ and $y$, we have $|\rho(G^{(i,j)}(x))-\rho(G^{(i,j)}(y))|$ to be upper bounded by $|\rho(G^{(i,j-1)}(x)) - \rho(G^{(i,j-1)}(y))|$ multiple to $\Delta |r^{(j-1)\rightarrow j}(x) - r^{(j-1)\rightarrow j}(y)|$.
Applying this repeatedly, for $l\leq j\leq L$, $|\rho(G^{(l,L)}(x))-\rho(G^{(l,L)}(y))|$ is bounded by
\begin{align}\label{Ineq-14}
\begin{split}
&|\rho(G^{(l,l+1)}(x))-\rho(G^{(l,l+1)}(y))| \\
  & \Delta_{k=l+2}^L |r^{(k-1)\rightarrow k}(x) - r^{(k-1)\rightarrow k}(y)| \\
= & \Delta_{k=l+1}^L |r^{(k-1)\rightarrow k}(x) - r^{(k-1)\rightarrow k}(y)|
\end{split}
\end{align}
Note that in GC-Net again the operation $\Delta$ is matrix concatenation. We use  the correlation $\rho$ on normalized filters \cref{eq:def-connectivity} as $r^{l\rightarrow (l+1)} (x)$ and the summand in \cref{Ineq-14} is
\begin{align}
\begin{split}
& |r^{(k-1)\rightarrow k}(x) - r^{(k-1)\rightarrow k}(y)| \\
= & |\mathbb{E}_{(X,Y)\sim D}[f^{(k-1)}(x)f^{(k)}(x)|Y] \\
  & - \mathbb{E}_{(X,Y)\sim D}[f^{(k-1)}(y)f^{(k)}(y)|Y]|,
\end{split}
\label{Ineq-15}
\end{align}
using Jensen's inequality,
\cref{Ineq-15} is upper bounded by 
\begin{equation}\label{Ineq-16}
 \mathbb{E}_{(X,Y)\sim D}[|f^{(k-1)}(x)f^{(k)}(x)-f^{(k-1)}(y)f^{(k)}(y)||Y].
\end{equation}
On the other hand using the definition of filters, we know that $f^{(k)}(x)=\sigma^{(k)}(\omega^{(k)}f^{(k-1)}(x)+b^{(k)})$. Therefore the upper bound \cref{Ineq-16} becomes
\begin{flalign*}
  & \hphantom{{}={}}\mathbb{E}_{(X,Y)\sim D}[|f^{(k-1)}(x)\sigma^{(k)}(\omega^{(k)}f^{(k-1)}(x)+b^{(k)}) & \\ 
  & \quad -f^{(k-1)}(y)\sigma^{(k)}(\omega^{(k)}f^{(k-1)}(y)+b^{(k)})||Y]. & \hspace{-2em} \tag{\theequation} \stepcounter{equation}
\end{flalign*}
\vspace{-2ex}
For simplicity, let $z_x=f^{(k-1)}(x)$ \& $z_y=f^{(k-1)}(y)$, then
\begin{align}
    &|z_x\sigma^{(k)}(\omega^{(k)}z_x+b^{(k)})-z_y\sigma^{(k)}(\omega^{(k)}z_y+b^{(k)})| \nonumber\\
   & = |z_x\sigma^{(k)}(\omega^{(k)}z_x+b^{(k)})-z_y\sigma^{(k)}(\omega^{(k)}z_x+b^{(k)})\nonumber\\
  &  \quad +z_y\sigma^{(k)}(\omega^{(k)}z_x+b^{(k)})-z_y\sigma^{(k)}(\omega^{(k)}z_y+b^{(k)})|\nonumber\\
  &  \leq |z_x-z_y||\sigma^{(k)}(\omega^{(k)}z_x+b^{(k)})|\label{Ineq-17}\\
   &+|z_y||\sigma^{(k)}(\omega^{(k)}z_x+b^{(k)})-\sigma^{(k)}(\omega^{(k)}z_y+b^{(k)})|. 
   \label{Ineq-18}
\end{align}
Since $\sigma^{(k)}$ is bounded, there exists a constant that $\sigma^{(k)}\leq C^{(k)}_\sigma$. Using Lipschitz continuous assumption for activation function and the filter value, $|z_x-z_y|$, are bounded,  
\begin{align}
\begin{split}
  |f^{(k-1)}(x) - f^{(k-1)}(y)| 
\leq & C^{(k-1)}|\omega^{(k-1)}||x-y|.
\end{split}
\label{Ineq.05}
\end{align}
Therefore, \cref{Ineq-17} and \cref{Ineq-18} are respectively bounded  
\begin{gather}
\begin{aligned}
|z_x-z_y||\sigma^{(k)}(&\omega^{(k)}z_x+b^{(k)})| \\
\leq & C^{(k)}_\sigma C^{(k-1)}|\omega^{(k-1)}||x-y|,
\end{aligned}
\label{Ineq-19} \\[1ex]
\begin{aligned}
&|\sigma^{(k)}(\omega^{(k)}z_x+b^{(k)})-\sigma^{(k)}(\omega^{(k)}z_y+b^{(k)})| \\
&\quad\leq C^{(k)}_\sigma |\omega^{(k)}||z_x-z_y|, \quad\rightarrow \text{using \cref{Ineq.05}} \\
&\quad\leq C^{(k-1)}C^{(k)}_\sigma |\omega^{(k)}||\omega^{(k-1)}||x-y|.
\end{aligned}
\label{Ineq-21}
\end{gather}
Applying upper bounds \cref{Ineq-19} and \cref{Ineq-21} and since $|z_y|$ is bounded by say $C_y$:
\begin{align}
\begin{split}
|z_x\sigma^{(k)}(&\omega^{(k)}z_x+b^{(k)}) - z_y\sigma^{(k)}(\omega^{(k)}z_y+b^{(k)})| \\
\leq C^{(k)}_\sigma & C^{(k-1)}|\omega^{(k-1)}||x-y| \\
& + C_y C^{(k-1)}C^{(k)}_\sigma |\omega^{(k)}||\omega^{(k-1)}||x-y| \\
= C^{(k)}_\sigma & C^{(k-1)}|\omega^{(k-1)}||x-y|(1 + C_y|\omega^{(k)}|) \\
\leq C^{(k-1,k)} & |\omega^{(k-1)}\omega^{(k)}||x-y|
\end{split}
\label{Ineq-11}
\end{align}
where $C^{(k-1,k)}=C C^{(k-1)}C^{(k)}_\sigma$. The last inequality in \cref{Ineq-11} holds true because  there exists a constant $C$ such that  $(1+C_y|\omega^{(k)}|\big)\leq C \; |\omega^{(k)}|$.
By
substituting $x=x^{(k-1)}_n$ and $y=\tilde{s}^{(l)} \odot x^{(k-1)}_n$ in \cref{Ineq-11} we have 
\begin{align}\label{Ineq.23}
 &C^{(k-1,k)}|\omega^{(k-1)}\omega^{(k)}| |x^{(k-1)}_n-\tilde{s}^{(k-1)} \odot x^{(k-1)}_n|\nonumber\\
 &\quad = C^{(k-1,k)}|\omega^{(k-1)}\omega^{(k)}| (\mathbf{1}-\tilde{s}^{(k-1)}) \odot |x^{(k-1)}_n|.
\end{align}
This upper bounds $|r^{(k-1)\rightarrow k}(x) -  r^{(k-1)\rightarrow k}(y)|$ by
\begin{align}
\begin{split}
C^{(k-1,k)}\mathbb{E}_{(X,Y)\sim D}[&|\omega^{(k-1)}\omega^{(k)}| \\
& \times (\mathbf{1}-\tilde{s}^{(k-1)}) \odot |x^{(k-1)}_n||Y].
\end{split}
\end{align}
From \cref{Ineq-14}, we bound $|\rho(G^{(l,L)}(x))-\rho(G^{(l,L)}(y))|$ by 
\begin{align}\label{Ineq-25}
\begin{split}
\Delta_{k=l+1}^L C^{(k-1,k)}\mathbb{E}_{(X,Y)\sim D}[&|\omega^{(k-1)}\omega^{(k)}| \\
\times (\mathbf{1}-\tilde{s}^{(k-1)}) \odot |x^{(k-1)}_n||Y].
\end{split}
\end{align}
Hence in \cref{eq2:objective}, the sample loss $\mathcal{F}$ is bounded by
\begin{align}
\begin{split}
\varphi( s^{(L)}, & \Delta_{k=l+1}^L C^{(k-1,k)} \times \mathbb{E}_{(X,Y)\sim D}[|\omega^{(k-1)}\omega^{(k)}| \\
& \quad \times (\mathbf{1}-\tilde{s}^{(l)}) \odot |x^{(k-1)}_n||Y]).
\end{split}
\end{align}
{\bf Assumption 1:} $\varphi$ function is satisfied in the 
inequality:
\begin{align}\label{assump.1}
&\varphi (z,\Delta_{k=i}^j \; u_k)\leq \;\Delta_{k=i}^j\varphi(z,u_k), \nonumber\\
&\Delta_{k=i}^j\varphi(z,\alpha \; u_k)\leq \alpha\; \Delta_{k=i}^j\varphi(z,u_k), \;\; \hbox{$\alpha$ is constant.}
\end{align}
Under Assum.~1, using~\cref{assump.1} 
the loss $\mathcal{F}$ is bounded by  
\begin{align}
& \Delta_{k=l+1}^L C^{(k-1,k)} \varphi( s^{(L)}, \quad |\omega^{(k-1)}\omega^{(k)}|\mathbb{E}_{(X,Y)\sim D}[ \nonumber \\
& (\mathbf{1}-\tilde{s}^{(l)}) \odot |x^{(k-1)}_n||Y]) = \Delta_{k=l+1}^L C^{(k-1,k)} \varphi( s^{(L)}, \nonumber \\
& |\omega^{(k-1)}\omega^{(k)}| (\mathbf{1}-\tilde{s}^{(l)})\odot \mathbb{E}_{(X,Y)\sim D}[|x^{(k-1)}_n||Y]).
\end{align}
Note that $|x^{(l)}_n|$ is bounded i.e. $|x^{(l)}_n|\leq C_{n}^{(l)}$, 
$\mathbb{E}_{(X,Y)\sim D}[ |x^{(k-1)}_n||Y]\leq C_n^{(k-1)}$:
\begin{align}
& \sum_{n = 1}^{N} \mathcal{F}(\tilde{s}^{(l)} \vert x^{(l)}_n, s^{(L)}; G^{(l+1,L)}) \leq \sum_{n = 1}^{N} \Delta_{k=l+1}^L C^{(k-1,k)} \nonumber \\
& C_n^{(k-1)}
 \varphi( s^{(L)}, |\omega^{(k-1)}\omega^{(k)}| (\mathbf{1}-\tilde{s}^{(l)})), \text{ equivalently} \nonumber \\
& \leq \Delta_{k=l+1}^L \tilde{C}^{(k-1,k)} \varphi( s^{(L)}, |\omega^{(k-1)}\omega^{(k)}| (\mathbf{1}-\tilde{s}^{(l)}))
\end{align}
where $\tilde{C}^{(k-1,k)}={C}^{(k-1,k)}\;\sum_{n = 1}^{N} C_n^{(k-1)}$. The objective function \cref{eq1:objective} becomes
\begin{equation}
\begin{split}
\mathop{argmin}\limits_{\tilde{s}^{(l)}} \Delta_{k=l+1}^L \tilde{C}^{(k-1,k)} \varphi\left( s^{(L)}, |\omega^{(k-1)}\omega^{(k)}| (\mathbf{1}-\tilde{s}^{(l)})\right)
\end{split}
\end{equation}
In GC-Net $F_{GC}$, $\varphi$ is dot product as follows
\begin{align}
\begin{split}
& \varphi( s^{(L)}, |\omega^{(k-1)}\omega^{(k)}| (\mathbf{1}-\tilde{s}^{(l)})) \\
= & \langle s^{(L)}, |\omega^{(k-1)}\omega^{(k)}| (\mathbf{1}-\tilde{s}^{(l)})\rangle \\
= & \langle (|\omega^{(k-1)}\omega^{(k)}|)^\intercal s^{(L)}, (\mathbf{1}-\tilde{s}^{(l)})\rangle=\sum_i g^{k-1,k}_i(1-\tilde{s}^{(l)}_i)
\end{split}
\end{align}
where $g^{k-1,k}= (|\omega^{(k-1)}\omega^{(k)}|)^\intercal s^{(L)}$.  This implies that \cref{eq1:objective} becomes
\begin{align}
\mathop{argmin}\limits_{\tilde{s}^{(l)}}\Delta_{k=l+1}^L \tilde{C}^{(k-1,k)} \sum_i g^{k-1,k}_i (1-\tilde{s}^{(l)}_i),
\end{align}
or equivalently the following maximization problem:
\begin{align}\label{sub.opt2}
 \mathop{argmax}\limits_{\tilde{s}^{(l)}} \Delta_{k=l+1}^L \;\tilde{C}^{(k-1,k)}\; \;\sum_i g^{k-1,k}_i \tilde{s}^{(l)}_i.
\end{align}
The optimal solution to~\cref{sub.opt2} is sub-optimal with respect
to the original objective in~\cref{eq1:objective}. The solution of~\cref{sub.opt2} captures the importance of filters (neurons) based on the connectivity $r^{l\rightarrow (l+1)}$ in GC-Net. In ~\cref{sub.opt2}, because $g^{k-1,k}= \big(|\omega^{(k-1)}\omega^{(k)}|\big)^\intercal s^{(L)}$, we can infer that there is a tight connection between GC-Net pruning and magnitude-based pruning, however, sparsifying $F_{GC}$ and mapping it back to the $F_O$ is equivalent to taking into account the $\omega^{(l)}\omega^{(l+1)}$ for all layers ($l=1,\ldots,L$) during pruning. This means that the weight multiplication is incorporated indirectly when applying GC-Net guidance whereas the magnitude pruning does not utilize layer-base connectivity.

\section{Discussion}\label{sec6}
Recent work has explored various approaches to improve network efficiency while maintaining and enhancing robustness. A Winning Hand \cite{diffenderfer2021winning} demonstrated that certain compression techniques, especially "lottery ticket-style" approaches, can inherently improve out-of-distribution robustness. DepGraph \cite{fang2023depgraph} introduced a generalized structural pruning method applicable across various network architectures. It explicitly models dependencies between layers to group coupled parameters for pruning, aiming for broad applicability rather than focusing specifically on DSs. In contrast, GC-Net takes a unique approach by introducing a companion network that guides pruning based on layer connectivity. This allows us to specifically target generalization under DSs while combining aspects of connectivity- and magnitude-based pruning methods.

\section{Conclusion}\label{sec7}
This study introduces GC-Net, a novel approach to address the DS challenge in sparse neural networks. Our experiments across various architectures, pruning methods, and datasets demonstrate the effectiveness of GC-Net in improving the robustness and adaptability of sparse models. Certain GC-Net hybrid models outperform or match traditionally pruned models across different sparsity levels and distribution shifts. Optimal GC-Net application varies with network architecture and pruning method, with notable benefits observed in later layers of VGG16-BN and specific portions of ResNet-18. 
While our study provides valuable theoretical and experimental insights into the potential of GC-Net, several avenues for future research remain, such as: 
\begin{itemize}
    \item \textbf{Dynamic Adaptation:} dynamically adjust GC-Net application during training or inference
    \item \textbf{Cross-Dataset Pruning:} investigate the effectiveness of pruning models using GC-Nets created from different datasets
    \item \textbf{Heterogeneous Architectures:} scenarios where the original network and GC-Net have different architectures
\end{itemize}
These opportunities aim to further enhance the robustness and flexibility of GC-Net in addressing distribution shift challenges across diverse complex scenarios.

\textbf{Acknowledgments:}
This work has been supported by the National Science Foundation (NSF) NSF CAREER-CCF 2451457 and Maine Space Grant Consortium (MSGC); the findings are those of the authors only and do not represent any position of these funding bodies.

\section{Appendix A}\label{secA1}
\subsection{Additional Experiments - Pearson Correlation}\label{subsec7}
Additional results for ResNet-18 and VGG16-BN for L1-norm, L2-norm, C-SNIP, and OS-SynFlow pruning using Pearson correlation GC-Net are included below.

All hybrid methods maintain competitive or superior performance at lower sparsity levels than the original model. Additionally, for CIFAR-10 and FMNIST the best performing hybrid methods appear to be GC-Net-BH, and GC-Net-B25\%, and all hybrid methods appear to maintain competitive accuracy at higher sparsity levels. The pruning results for for the OS-SynFlow and C-SNIP pruning indicate an overall benefit for applying the GC-Net method to the later layers of the network. 

\cref{tab:performance_comparison_resnet18_OS_SynFlow_20,tab:performance_comparison_resnet18_OS_SynFlow_40,tab:performance_comparison_resnet18_OS_SynFlow_60,tab:performance_comparison_resnet18_OS_SynFlow_80} indicate a pattern in the performance of the hybrid models, as GC-Net-B25\% was frequently the best performing hybrid model, particularly for performance on the CIFAR-10 distribution datasets at 20\% sparsity. The hybrid models showed a competitive or superior performance at higher sparsities for the FMNIST trials, particularly at 60\% sparsity where a hybrid model outperformed the original model on all data variations and matched the performance on the original dataset. For Tiny-IN, there is also an improved performance of GC-Net-FH in comparison to other hybrid models, and at 80\% sparsity the hybrid models outperformed the original model for all dataset variations. Also, it appears the hybrid models often performed best on the RNB dataset variation for CIFAR-10, the LO variation for FMNIST, and CJG and LO for Tiny-IN. The pattern in the best performing hybrid model, combined with the variations in the best performing distribution shift variation indicates that networks benefit from having the later portion of the network pruned using connectivity, rather than the earlier layers. Additionally, while the GC-Net method improves DS accuracy, the type of DS GC-Net is most beneficial for may be dependent on the original dataset.

\begin{table}[!ht]
\centering
\setlength{\tabcolsep}{4pt} 
\footnotesize
\begin{tabular}{|>{\centering\arraybackslash}p{1.8cm}|c|cccc|}
\hline
\multirow{2}{*}{\makecell[c]{Architecture}} & \multirow{2}{*}{Hybrid Type} & \multicolumn{4}{c|}{\textbf{L1-Norm - Sparsity 20\%}} \\
 &  & Orig($Acc_1$) & CJG & RNB & LO \\
\hline
\multirow{5}{*}{\makecell[c]{ResNet-18\\(CIFAR-10)\\$Acc_O$: 93.14\%}} & Full GC-Net & 92.25 & 71.69 & 22.76 & 13.82 \\
& GC-Net-FH & 92.21 & 71.51 & 22.26 & \textbf{13.95} \\
& GC-Net-BH & 93.05 & 73.52 & 23.63 & 13.82 \\
& GC-Net-B25\% & \textbf{93.11} & 73.32 & 24.24 & 13.75 \\
& Original & \cellcolor{yellow!25}93.04 & \cellcolor{yellow!25}\textbf{74.38} & \cellcolor{yellow!25}\textbf{24.49} & \cellcolor{yellow!25}13.69 \\
\hline
\multirow{5}{*}{\makecell[c]{ResNet-18\\(FMNIST)\\$Acc_O$: 93.63\%}} & Full GC-Net & 92.61 & 27.27 & 44.62 & 20.4 \\
& GC-Net-FH & 92.97 & \textbf{29.49} & \textbf{56.1} & \textbf{23.56} \\
& GC-Net-BH & 93.15 & 26.72 & 50.35 & 20.64 \\
& GC-Net-B25\% & 93.34 & 26.2 & 52.09 & 21.56 \\
& Original & \cellcolor{yellow!25}\textbf{93.41} & \cellcolor{yellow!25}26.65 & \cellcolor{yellow!25}55.71 & \cellcolor{yellow!25}22.68 \\
\hline
\multirow{5}{*}{\makecell[c]{ResNet-18\\(Tiny-IN)\\$Acc_O$: 65.15\%}} & Full GC-Net & 61.37 & 6.46 & 4.98 & 5.16 \\
& GC-Net-FH & 64.08 & 7.23 & 5.05 & \textbf{5.85} \\
& GC-Net-BH & 62.65 & 6.27 & 4.59 & 4.74 \\
& GC-Net-B25\% & 62.79 & 6.62 & 4.74 & 5.03 \\
& Original & \cellcolor{yellow!25}\textbf{64.5} & \cellcolor{yellow!25}\textbf{7.29} & \cellcolor{yellow!25}\textbf{5.3} & \cellcolor{yellow!25}5.38 \\
\hline
\end{tabular}
\caption{ResNet-18 \textbf{L1-Pruning} average accuracy (\%) across datasets and hybrid types at \textbf{20\% sparsity} using Pearson correlation GC-Net. Results for applying the pruning method directly to the original model, without a hybrid method, are highlighted in yellow.}
\label{tab:performance_comparison_resnet18_L1_20}
\end{table}

\begin{table}[!ht]
\centering
\setlength{\tabcolsep}{4pt}
\footnotesize
\begin{tabular}{|>{\centering\arraybackslash}p{1.8cm}|c|cccc|}
\hline
\multirow{2}{*}{\makecell[c]{Architecture}} & \multirow{2}{*}{Hybrid Type} & \multicolumn{4}{c|}{\textbf{L1-Norm - Sparsity 40\%}} \\
 &  & Orig($Acc_1$) & CJG & RNB & LO \\
\hline
\multirow{5}{*}{\makecell[c]{ResNet-18\\(CIFAR-10)\\$Acc_O$: 93.14\%}} & Full GC-Net & 90.93 & 70.24 & 22.24 & 13.78 \\
& GC-Net-FH & 91.12 & 69.76 & 22.18 & 13.47 \\
& GC-Net-BH & 92.85 & 73.43 & 24.32 & 13.87 \\
& GC-Net-B25\% & \textbf{93.08} & 73.73 & \textbf{24.77} & 13.88 \\
& Original & \cellcolor{yellow!25}93.07 & \cellcolor{yellow!25}\textbf{73.83} & \cellcolor{yellow!25}24.43 & \cellcolor{yellow!25}\textbf{14.19} \\
\hline
\multirow{5}{*}{\makecell[c]{ResNet-18\\(FMNIST)\\$Acc_O$: 93.63\%}} & Full GC-Net & 92.52 & 25.57 & 39.25 & 21.27 \\
& GC-Net-FH & 92.64 & 27.31 & 51.46 & 23.0 \\
& GC-Net-BH & 92.91 & \textbf{28.94} & 49.34 & 21.95 \\
& GC-Net-B25\% & 93.14 & 26.76 & 50.49 & 21.36 \\
& Original & \cellcolor{yellow!25}\textbf{93.38} & \cellcolor{yellow!25}26.57 & \cellcolor{yellow!25}\textbf{56.02} & \cellcolor{yellow!25}\textbf{21.91} \\
\hline
\multirow{5}{*}{\makecell[c]{ResNet-18\\(Tiny-IN)\\$Acc_O$: 65.15\%}} & Full GC-Net & 56.8 & 6.34 & 3.76 & 4.76 \\
& GC-Net-FH & 62.48 & 7.08 & 4.8 & \textbf{5.77} \\
& GC-Net-BH & 60.09 & 6.74 & 4.04 & 5.01 \\
& GC-Net-B25\% & 61.43 & 6.52 & 4.64 & 5.34 \\
& Original & \cellcolor{yellow!25}\textbf{64.74} & \cellcolor{yellow!25}\textbf{7.12} & \cellcolor{yellow!25}\textbf{5.05} & \cellcolor{yellow!25}5.57 \\
\hline
\end{tabular}
\caption{ResNet-18 \textbf{L1-Pruning} average accuracy (\%) across datasets and hybrid types at \textbf{40\% sparsity} using Pearson correlation GC-Net. Results for applying the pruning method directly to the original model, without a hybrid method, are highlighted in yellow.}
\label{tab:performance_comparison_resnet18_L1_40}
\end{table}

\begin{table}[!ht]
\centering
\setlength{\tabcolsep}{4pt}
\footnotesize
\begin{tabular}{|>{\centering\arraybackslash}p{1.8cm}|c|cccc|}
\hline
\multirow{2}{*}{\makecell[c]{Architecture}} & \multirow{2}{*}{Hybrid Type} & \multicolumn{4}{c|}{\textbf{L1-Norm - Sparsity 60\%}} \\
 &  & Orig($Acc_1$) & CJG & RNB & LO \\
\hline
\multirow{5}{*}{\makecell[c]{ResNet-18\\(CIFAR-10)\\$Acc_O$: 93.14\%}} & Full GC-Net & 89.22 & 67.95 & 21.45 & 13.7 \\
& GC-Net-FH & 89.51 & 67.93 & 21.85 & 13.47 \\
& GC-Net-BH & 92.34 & 72.04 & 22.03 & 13.37 \\
& GC-Net-B25\% & \textbf{92.66} & \textbf{73.13} & \textbf{23.42} & 13.6 \\
& Original & \cellcolor{yellow!25}92.6 & \cellcolor{yellow!25}72.89 & \cellcolor{yellow!25}23.29 & \cellcolor{yellow!25}\textbf{13.78} \\
\hline
\multirow{5}{*}{\makecell[c]{ResNet-18\\(FMNIST)\\$Acc_O$: 93.63\%}} & Full GC-Net & 91.46 & 17.19 & 30.64 & 16.77 \\
& GC-Net-FH & 92.19 & 17.32 & 26.31 & 15.8 \\
& GC-Net-BH & 92.83 & \textbf{28.75} & \textbf{57.51} & \textbf{24.29} \\
& GC-Net-B25\% & 92.84 & 28.25 & 46.03 & 21.06 \\
& Original & \cellcolor{yellow!25}\textbf{93.43} & \cellcolor{yellow!25}27.37 & \cellcolor{yellow!25}57.23 & \cellcolor{yellow!25}22.36 \\
\hline
\multirow{5}{*}{\makecell[c]{ResNet-18\\(Tiny-IN)\\$Acc_O$: 65.15\%}} & Full GC-Net & 48.93 & 5.77 & 2.98 & 4.27 \\
& GC-Net-FH & 60.45 & 7.15 & 3.67 & \textbf{5.95} \\
& GC-Net-BH & 54.9 & 6.26 & 3.85 & 5.03 \\
& GC-Net-B25\% & 58.61 & 6.74 & 4.27 & 5.53 \\
& Original & \cellcolor{yellow!25}\textbf{64.44} & \cellcolor{yellow!25}\textbf{7.34} & \cellcolor{yellow!25}\textbf{4.8} & \cellcolor{yellow!25}5.91 \\
\hline
\end{tabular}
\caption{ResNet-18 \textbf{L1-Pruning} average accuracy (\%) across datasets and hybrid types at \textbf{60\% sparsity} using Pearson correlation GC-Net. Results for applying the pruning method directly to the original model, without a hybrid method, are highlighted in yellow.}
\label{tab:performance_comparison_resnet18_L1_60}
\end{table}

\begin{table}[!ht]
\centering
\setlength{\tabcolsep}{4pt}
\footnotesize
\begin{tabular}{|>{\centering\arraybackslash}p{1.8cm}|c|cccc|}
\hline
\multirow{2}{*}{\makecell[c]{Architecture}} & \multirow{2}{*}{Hybrid Type} & \multicolumn{4}{c|}{\textbf{L1-Norm - Sparsity 80\%}} \\
 &  & Orig($Acc_1$) & CJG & RNB & LO \\
\hline
\multirow{5}{*}{\makecell[c]{ResNet-18\\(CIFAR-10)\\$Acc_O$: 93.14\%}} & Full GC-Net & 57.01 & 41.24 & 17.12 & 11.96 \\
& GC-Net-FH & 67.76 & 48.32 & 18.03 & 13.11 \\
& GC-Net-BH & 89.84 & 69.89 & 21.74 & \textbf{13.37} \\
& GC-Net-B25\% & \textbf{91.05} & 71.52 & 23.34 & 13.35 \\
& Original & \cellcolor{yellow!25}90.91 & \cellcolor{yellow!25}\textbf{71.71} & \cellcolor{yellow!25}\textbf{23.8} & \cellcolor{yellow!25}13.19 \\
\hline
\multirow{5}{*}{\makecell[c]{ResNet-18\\(FMNIST)\\$Acc_O$: 93.63\%}} & Full GC-Net & 90.53 & 23.91 & 35.01 & 17.91 \\
& GC-Net-FH & 90.57 & 20.23 & 32.37 & 18.58 \\
& GC-Net-BH & 91.96 & 30.11 & 54.5 & \textbf{22.77} \\
& GC-Net-B25\% & 92.3 & \textbf{32.48} & 59.53 & 22.24 \\
& Original & \cellcolor{yellow!25}\textbf{92.73} & \cellcolor{yellow!25}29.65 & \cellcolor{yellow!25}\textbf{60.33} & \cellcolor{yellow!25}22.06 \\
\hline
\multirow{5}{*}{\makecell[c]{ResNet-18\\(Tiny-IN)\\$Acc_O$: 65.15\%}} & Full GC-Net & 33.6 & 2.65 & 1.28 & 2.01 \\
& GC-Net-FH & 52.24 & 5.97 & 2.82 & 5.4 \\
& GC-Net-BH & 45.01 & 5.16 & 2.47 & 4.02 \\
& GC-Net-B25\% & 52.19 & 6.17 & 3.8 & 4.74 \\
& Original & \cellcolor{yellow!25}\textbf{60.14} & \cellcolor{yellow!25}\textbf{7.05} & \cellcolor{yellow!25}\textbf{4.32} & \cellcolor{yellow!25}\textbf{5.74} \\
\hline
\end{tabular}
\caption{ResNet-18 \textbf{L1-Pruning} average accuracy (\%) across datasets and hybrid types at \textbf{80\% sparsity} using Pearson correlation GC-Net. Results for applying the pruning method directly to the original model, without a hybrid method, are highlighted in yellow.}
\label{tab:performance_comparison_resnet18_L1_80}
\end{table}

\begin{table}[!ht]
\centering
\setlength{\tabcolsep}{4pt}
\footnotesize
\begin{tabular}{|>{\centering\arraybackslash}p{1.8cm}|c|cccc|}
\hline
\multirow{2}{*}{\makecell[c]{Architecture}} & \multirow{2}{*}{Hybrid Type} & \multicolumn{4}{c|}{\textbf{L2-Norm - Sparsity 20\%}} \\
 &  & Orig($Acc_1$) & CJG & RNB & LO \\
\hline
\multirow{5}{*}{\makecell[c]{ResNet-18\\(CIFAR-10)\\$Acc_O$: 93.14\%}} & Full GC-Net & 91.27 & 69.77 & 24.2 & \textbf{14.33} \\
& GC-Net-FH & 91.62 & 71.12 & \textbf{24.83} & 13.39 \\
& GC-Net-BH & 91.76 & 70.89 & 23.23 & 13.74 \\
& GC-Net-B25\% & 91.85 & 70.63 & 23.46 & 14.15 \\
& Original & \cellcolor{yellow!25}\textbf{91.99} & \cellcolor{yellow!25}\textbf{71.62} & \cellcolor{yellow!25}24.3 & \cellcolor{yellow!25}13.75 \\
\hline
\multirow{5}{*}{\makecell[c]{ResNet-18\\(FMNIST)\\$Acc_O$: 93.63\%}} & Full GC-Net & 92.75 & 27.74 & 43.37 & 21.61 \\
& GC-Net-FH & 92.88 & 25.83 & 38.93 & 20.62 \\
& GC-Net-BH & \textbf{92.96} & \textbf{30.16} & \textbf{63.27} & \textbf{25.17} \\
& GC-Net-B25\% & 92.91 & 28.87 & 55.93 & 22.19 \\
& Original & \cellcolor{yellow!25}92.93 & \cellcolor{yellow!25}28.1 & \cellcolor{yellow!25}53.63 & \cellcolor{yellow!25}22.72 \\
\hline
\multirow{5}{*}{\makecell[c]{ResNet-18\\(Tiny-IN)\\$Acc_O$: 65.15\%}} & Full GC-Net & 58.65 & 6.68 & 3.66 & 5.25 \\
& GC-Net-FH & 60.04 & \textbf{7.82} & 3.78 & \textbf{5.84} \\
& GC-Net-BH & 59.91 & 6.3 & 3.38 & 4.77 \\
& GC-Net-B25\% & 60.62 & 7.24 & \textbf{3.97} & 5.35 \\
& Original & \cellcolor{yellow!25}\textbf{61.18} & \cellcolor{yellow!25}7.13 & \cellcolor{yellow!25}3.58 & \cellcolor{yellow!25}5.48 \\
\hline
\end{tabular}
\caption{ResNet-18 \textbf{L2-Pruning} average accuracy (\%) across datasets and hybrid types at \textbf{20\% sparsity} using Pearson correlation GC-Net. Results for applying the pruning method directly to the original model, without a hybrid method, are highlighted in yellow.}
\label{tab:performance_comparison_resnet18_L2_20}
\end{table}

\begin{table}[!ht]
\centering
\setlength{\tabcolsep}{4pt}
\footnotesize
\begin{tabular}{|>{\centering\arraybackslash}p{1.8cm}|c|cccc|}
\hline
\multirow{2}{*}{\makecell[c]{Architecture}} & \multirow{2}{*}{Hybrid Type} & \multicolumn{4}{c|}{\textbf{L2-Norm - Sparsity 40\%}} \\
 &  & Orig($Acc_1$) & CJG & RNB & LO \\
\hline
\multirow{5}{*}{\makecell[c]{ResNet-18\\(CIFAR-10)\\$Acc_O$: 93.14\%}} & Full GC-Net & 86.43 & 59.38 & 21.88 & 14.21 \\
& GC-Net-FH & 89.56 & 67.99 & \textbf{24.92} & 13.99 \\
& GC-Net-BH & 84.22 & 60.45 & 18.69 & \textbf{14.41} \\
& GC-Net-B25\% & 86.18 & 62.97 & 18.8 & 14.02 \\
& Original & \cellcolor{yellow!25}\textbf{90.77} & \cellcolor{yellow!25}\textbf{69.51} & \cellcolor{yellow!25}21.61 & \cellcolor{yellow!25}14.01 \\
\hline
\multirow{5}{*}{\makecell[c]{ResNet-18\\(FMNIST)\\$Acc_O$: 93.63\%}} & Full GC-Net & 92.01 & 27.07 & 53.34 & 24.31 \\
& GC-Net-FH & 91.49 & 28.3 & 44.85 & 23.58 \\
& GC-Net-BH & 92.48 & 30.18 & \textbf{61.36} & 25.19 \\
& GC-Net-B25\% & 92.31 & 31.71 & 59.75 & 27.53 \\
& Original & \cellcolor{yellow!25}\textbf{92.54} & \cellcolor{yellow!25}\textbf{32.52} & \cellcolor{yellow!25}61.35 & \cellcolor{yellow!25}\textbf{28.88} \\
\hline
\multirow{5}{*}{\makecell[c]{ResNet-18\\(Tiny-IN)\\$Acc_O$: 65.15\%}} & Full GC-Net & 48.7 & 5.45 & 2.22 & 3.85 \\
& GC-Net-FH & 52.13 & \textbf{6.21} & \textbf{3.43} & \textbf{5.2} \\
& GC-Net-BH & 50.82 & 5.26 & 2.57 & 3.51 \\
& GC-Net-B25\% & 52.7 & 5.8 & 2.81 & 4.57 \\
& Original & \cellcolor{yellow!25}\textbf{53.84} & \cellcolor{yellow!25}6.15 & \cellcolor{yellow!25}2.6 & \cellcolor{yellow!25}4.67 \\
\hline
\end{tabular}
\caption{ResNet-18 \textbf{L2-Pruning} average accuracy (\%) across datasets and hybrid types at \textbf{40\% sparsity} using Pearson correlation GC-Net. Results for applying the pruning method directly to the original model, without a hybrid method, are highlighted in yellow.}
\label{tab:performance_comparison_resnet18_L2_40}
\end{table}

\begin{table}[!ht]
\centering
\setlength{\tabcolsep}{4pt}
\footnotesize
\begin{tabular}{|>{\centering\arraybackslash}p{1.8cm}|c|cccc|}
\hline
\multirow{2}{*}{\makecell[c]{Architecture}} & \multirow{2}{*}{Hybrid Type} & \multicolumn{4}{c|}{\textbf{L2-Norm - Sparsity 60\%}} \\
 &  & Orig($Acc_1$) & CJG & RNB & LO \\
\hline
\multirow{5}{*}{\makecell[c]{ResNet-18\\(CIFAR-10)\\$Acc_O$: 93.14\%}} & Full GC-Net & 29.28 & 23.34 & 14.26 & 10.51 \\
& GC-Net-FH & 84.47 & 59.36 & \textbf{23.33} & \textbf{15.04} \\
& GC-Net-BH & 35.0 & 25.98 & 18.74 & 11.42 \\
& GC-Net-B25\% & 57.97 & 42.5 & 19.26 & 13.59 \\
& Original & \cellcolor{yellow!25}\textbf{87.43} & \cellcolor{yellow!25}\textbf{63.39} & \cellcolor{yellow!25}21.85 & \cellcolor{yellow!25}14.69 \\
\hline
\multirow{5}{*}{\makecell[c]{ResNet-18\\(FMNIST)\\$Acc_O$: 93.63\%}} & Full GC-Net & 91.32 & 21.74 & 32.36 & 18.19 \\
& GC-Net-FH & 91.38 & 23.57 & 35.62 & 18.15 \\
& GC-Net-BH & \textbf{91.67} & \textbf{30.89} & \textbf{58.53} & \textbf{30.27} \\
& GC-Net-B25\% & 91.65 & 22.26 & 31.53 & 20.29 \\
& Original & \cellcolor{yellow!25}91.66 & \cellcolor{yellow!25}24.83 & \cellcolor{yellow!25}45.6 & \cellcolor{yellow!25}22.86 \\
\hline
\multirow{5}{*}{\makecell[c]{ResNet-18\\(Tiny-IN)\\$Acc_O$: 65.15\%}} & Full GC-Net & 28.13 & 2.31 & 1.13 & 1.89 \\
& GC-Net-FH & 31.61 & 3.21 & 1.42 & 2.47 \\
& GC-Net-BH & 32.8 & 2.96 & 1.55 & 1.69 \\
& GC-Net-B25\% & 35.26 & 4.02 & 2.18 & 2.7 \\
& Original & \cellcolor{yellow!25}\textbf{38.32} & \cellcolor{yellow!25}\textbf{4.56} & \cellcolor{yellow!25}\textbf{2.2} & \cellcolor{yellow!25}\textbf{2.93} \\
\hline
\end{tabular}
\caption{ResNet-18 \textbf{L2-Pruning} average accuracy (\%) across datasets and hybrid types at \textbf{60\% sparsity} using Pearson correlation GC-Net. Results for applying the pruning method directly to the original model, without a hybrid method, are highlighted in yellow.}
\label{tab:performance_comparison_resnet18_L2_60}
\end{table}

\begin{table}[!ht]
\centering
\setlength{\tabcolsep}{4pt}
\footnotesize
\begin{tabular}{|>{\centering\arraybackslash}p{1.8cm}|c|cccc|}
\hline
\multirow{2}{*}{\makecell[c]{Architecture}} & \multirow{2}{*}{Hybrid Type} & \multicolumn{4}{c|}{\textbf{L2-Norm - Sparsity 80\%}} \\
 &  & Orig($Acc_1$) & CJG & RNB & LO \\
\hline
\multirow{5}{*}{\makecell[c]{ResNet-18\\(CIFAR-10)\\$Acc_O$: 93.14\%}} & Full GC-Net & 10.0 & 10.0 & 10.0 & 10.0 \\
& GC-Net-FH & 10.01 & 9.97 & 10.0 & 10.0 \\
& GC-Net-BH & 10.0 & 10.01 & 10.0 & 10.0 \\
& GC-Net-B25\% & 13.85 & 12.2 & 11.41 & 10.51 \\
& Original & \cellcolor{yellow!25}\textbf{31.2} & \cellcolor{yellow!25}\textbf{24.72} & \cellcolor{yellow!25}\textbf{14.93} & \cellcolor{yellow!25}\textbf{11.68} \\
\hline
\multirow{5}{*}{\makecell[c]{ResNet-18\\(FMNIST)\\$Acc_O$: 93.63\%}} & Full GC-Net & 90.74 & 17.07 & 23.1 & 17.23 \\
& GC-Net-FH & 90.38 & 19.99 & 24.99 & 17.52 \\
& GC-Net-BH & \textbf{90.64} & \textbf{23.37} & \textbf{41.88} & \textbf{26.45} \\
& GC-Net-B25\% & 90.26 & 22.41 & 41.62 & 20.91 \\
& Original & \cellcolor{yellow!25}90.43 & \cellcolor{yellow!25}13.78 & \cellcolor{yellow!25}39.35 & \cellcolor{yellow!25}18.55 \\
\hline
\multirow{5}{*}{\makecell[c]{ResNet-18\\(Tiny-IN)\\$Acc_O$: 65.15\%}} & Full GC-Net & 10.48 & 1.26 & 1.04 & 1.38 \\
& GC-Net-FH & 16.03 & 1.55 & 1.26 & 1.47 \\
& GC-Net-BH & 12.1 & 2.0 & 0.9 & 2.1 \\
& GC-Net-B25\% & 12.08 & 1.55 & 0.92 & 1.55 \\
& Original & \cellcolor{yellow!25}\textbf{16.68} & \cellcolor{yellow!25}\textbf{2.05} & \cellcolor{yellow!25}\textbf{1.38} & \cellcolor{yellow!25}\textbf{1.81} \\
\hline
\end{tabular}
\caption{ResNet-18 \textbf{L2-Pruning} average accuracy (\%) across datasets and hybrid types at \textbf{80\% sparsity} using Pearson correlation GC-Net. Results for applying the pruning method directly to the original model, without a hybrid method, are highlighted in yellow.}
\label{tab:performance_comparison_resnet18_L2_80}
\end{table}

\begin{table}[!ht]
\centering
\setlength{\tabcolsep}{4pt}
\footnotesize
\begin{tabular}{|>{\centering\arraybackslash}p{1.8cm}|c|cccc|}
\hline
\multirow{2}{*}{\makecell[c]{Architecture}} & \multirow{2}{*}{Hybrid Type} & \multicolumn{4}{c|}{\textbf{OS-SynFlow - Sparsity 20\%}} \\
 &  & Orig($Acc_1$) & CJG & RNB & LO \\
\hline
\multirow{5}{*}{\makecell[c]{ResNet-18\\(CIFAR-10)\\$Acc_O$: 93.14\%}} & Full GC-Net & 92.39 & 71.67 & 22.11 & 13.6 \\
& GC-Net-FH & 92.33 & 72.42 & 22.9 & 13.9 \\
& GC-Net-BH & 93.1 & 73.41 & 24.43 & 13.65 \\
& GC-Net-B25\% & 93.09 & \textbf{74.16} & \textbf{24.67} & \textbf{13.92} \\
& Original & \cellcolor{yellow!25}\textbf{93.15} & \cellcolor{yellow!25}73.87 & \cellcolor{yellow!25}24.48 & \cellcolor{yellow!25}13.32 \\
\hline
\multirow{5}{*}{\makecell[c]{ResNet-18\\(FMNIST)\\$Acc_O$: 93.63\%}} & Full GC-Net & 93.28 & \textbf{28.84} & 56.05 & \textbf{22.8} \\
& GC-Net-FH & 93.33 & 27.48 & \textbf{59.15} & 21.26 \\
& GC-Net-BH & 93.1 & 26.97 & 55.7 & 22.38 \\
& GC-Net-B25\% & \textbf{93.41} & 25.86 & 48.69 & 21.72 \\
& Original & \cellcolor{yellow!25}93.25 & \cellcolor{yellow!25}24.92 & \cellcolor{yellow!25}52.48 & \cellcolor{yellow!25}21.05 \\
\hline
\multirow{5}{*}{\makecell[c]{ResNet-18\\(Tiny-IN)\\$Acc_O$: 65.15\%}} & Full GC-Net & 63.32 & \textbf{7.3} & 5.05 & 5.48 \\
& GC-Net-FH & 60.69 & 7.14 & 4.84 & \textbf{5.81} \\
& GC-Net-BH & 63.74 & 6.77 & 4.83 & 5.4 \\
& GC-Net-B25\% & 63.9 & 7.12 & \textbf{5.17} & 5.5 \\
& Original & \cellcolor{yellow!25}\textbf{64.39} & \cellcolor{yellow!25}7.07 & \cellcolor{yellow!25}4.8 & \cellcolor{yellow!25}5.59 \\
\hline
\end{tabular}
\caption{ResNet-18 \textbf{OS-SynFlow} average accuracy (\%) across datasets and hybrid types at \textbf{20\% sparsity} using Pearson correlation. Results for applying the pruning method directly to the original model, without a hybrid method, are highlighted in yellow.}
\label{tab:performance_comparison_resnet18_OS_SynFlow_20}
\end{table}

\begin{table}[!ht]
\centering
\setlength{\tabcolsep}{4pt}
\footnotesize
\begin{tabular}{|>{\centering\arraybackslash}p{1.8cm}|c|cccc|}
\hline
\multirow{2}{*}{\makecell[c]{Architecture}} & \multirow{2}{*}{Hybrid Type} & \multicolumn{4}{c|}{\textbf{OS-SynFlow - Sparsity 40\%}} \\
 &  & Orig($Acc_1$) & CJG & RNB & LO \\
\hline
\multirow{5}{*}{\makecell[c]{ResNet-18\\(CIFAR-10)\\$Acc_O$: 93.14\%}} & Full GC-Net & 91.36 & 70.63 & 22.26 & 13.56 \\
& GC-Net-FH & 91.65 & 70.15 & 22.84 & \textbf{13.74} \\
& GC-Net-BH & 92.59 & 73.69 & 23.97 & 13.67 \\
& GC-Net-B25\% & \textbf{92.9} & \textbf{74.13} & 23.96 & 13.64 \\
& Original & \cellcolor{yellow!25}92.8 & \cellcolor{yellow!25}73.78 & \cellcolor{yellow!25}\textbf{24.17} & \cellcolor{yellow!25}13.58 \\
\hline
\multirow{5}{*}{\makecell[c]{ResNet-18\\(FMNIST)\\$Acc_O$: 93.63\%}} & Full GC-Net & 92.29 & 27.6 & 50.3 & 23.2 \\
& GC-Net-FH & 92.89 & 28.35 & 54.83 & \textbf{23.96} \\
& GC-Net-BH & 92.98 & 27.91 & 54.2 & 20.66 \\
& GC-Net-B25\% & 92.99 & \textbf{28.99} & \textbf{57.22} & 22.33 \\
& Original & \cellcolor{yellow!25}\textbf{93.43} & \cellcolor{yellow!25}25.54 & \cellcolor{yellow!25}52.69 & \cellcolor{yellow!25}21.36 \\
\hline
\multirow{5}{*}{\makecell[c]{ResNet-18\\(Tiny-IN)\\$Acc_O$: 65.15\%}} & Full GC-Net & 60.01 & 6.69 & 4.11 & 5.51 \\
& GC-Net-FH & 63.28 & \textbf{7.53} & 4.72 & \textbf{5.97} \\
& GC-Net-BH & 61.41 & 7.07 & 4.25 & 5.17 \\
& GC-Net-B25\% & 62.68 & 7.14 & \textbf{4.82} & 5.47 \\
& Original & \cellcolor{yellow!25}\textbf{64.25} & \cellcolor{yellow!25}7.5 & \cellcolor{yellow!25}4.5 & \cellcolor{yellow!25}5.49 \\
\hline
\end{tabular}
\caption{ResNet-18 \textbf{OS-SynFlow} average accuracy (\%) across datasets and hybrid types at \textbf{40\% sparsity} using Pearson correlation. Results for applying the pruning method directly to the original model, without a hybrid method, are highlighted in yellow.}
\label{tab:performance_comparison_resnet18_OS_SynFlow_40}
\end{table}

\begin{table}[!ht]
\centering
\setlength{\tabcolsep}{4pt}
\footnotesize
\begin{tabular}{|>{\centering\arraybackslash}p{1.8cm}|c|cccc|}
\hline
\multirow{2}{*}{\makecell[c]{Architecture}} & \multirow{2}{*}{Hybrid Type} & \multicolumn{4}{c|}{\textbf{OS-SynFlow - Sparsity 60\%}} \\
 &  & Orig($Acc_1$) & CJG & RNB & LO \\
\hline
\multirow{5}{*}{\makecell[c]{ResNet-18\\(CIFAR-10)\\$Acc_O$: 93.14\%}} & Full GC-Net & 89.23 & 67.01 & 23.71 & 13.62 \\
& GC-Net-FH & 89.98 & 47.49 & \textbf{24.03} & \textbf{14.15} \\
& GC-Net-BH & 92.19 & 72.6 & 22.66 & 13.36 \\
& GC-Net-B25\% & \textbf{92.41} & 72.94 & 23.3 & 13.77 \\
& Original & \cellcolor{yellow!25}92.36 & \cellcolor{yellow!25}\textbf{73.17} & \cellcolor{yellow!25}23.01 & \cellcolor{yellow!25}13.54 \\
\hline
\multirow{5}{*}{\makecell[c]{ResNet-18\\(FMNIST)\\$Acc_O$: 93.63\%}} & Full GC-Net & 91.53 & 22.54 & 37.85 & 18.31 \\
& GC-Net-FH & 92.08 & 23.51 & 41.63 & 19.97 \\
& GC-Net-BH & 92.78 & 29.24 & 56.35 & \textbf{22.74} \\
& GC-Net-B25\% & \textbf{92.89} & \textbf{29.54} & \textbf{59.32} & 22.63 \\
& Original & \cellcolor{yellow!25}\textbf{92.89} & \cellcolor{yellow!25}25.98 & \cellcolor{yellow!25}52.89 & \cellcolor{yellow!25}21.45 \\
\hline
\multirow{5}{*}{\makecell[c]{ResNet-18\\(Tiny-IN)\\$Acc_O$: 65.15\%}} & Full GC-Net & 51.47 & 5.42 & 2.96 & 4.7 \\
& GC-Net-FH & 59.83 & 7.03 & 3.5 & \textbf{6.13} \\
& GC-Net-BH & 56.96 & 6.59 & 3.95 & 5.3 \\
& GC-Net-B25\% & 60.4 & 6.83 & \textbf{4.11} & 5.58 \\
& Original & \cellcolor{yellow!25}\textbf{63.35} & \cellcolor{yellow!25}\textbf{7.35} & \cellcolor{yellow!25}4.06 & \cellcolor{yellow!25}5.92 \\
\hline
\end{tabular}
\caption{ResNet-18 \textbf{OS-SynFlow} average accuracy (\%) across datasets and hybrid types at \textbf{60\% sparsity} using Pearson correlation. Results for applying the pruning method directly to the original model, without a hybrid method, are highlighted in yellow.}
\label{tab:performance_comparison_resnet18_OS_SynFlow_60}
\end{table}

\begin{table}[!ht]
\centering
\setlength{\tabcolsep}{4pt}
\footnotesize
\begin{tabular}{|>{\centering\arraybackslash}p{1.8cm}|c|cccc|}
\hline
\multirow{2}{*}{\makecell[c]{Architecture}} & \multirow{2}{*}{Hybrid Type} & \multicolumn{4}{c|}{\textbf{OS-SynFlow - Sparsity 80\%}} \\
 &  & Orig($Acc_1$) & CJG & RNB & LO \\
\hline
\multirow{5}{*}{\makecell[c]{ResNet-18\\(CIFAR-10)\\$Acc_O$: 93.14\%}} & Full GC-Net & 26.29 & 22.34 & 13.67 & 10.62 \\
& GC-Net-FH & 66.74 & 46.02 & 18.2 & 12.64 \\
& GC-Net-BH & 88.33 & 67.68 & 23.22 & 13.81 \\
& GC-Net-B25\% & 90.35 & 69.9 & 24.65 & \textbf{14.38} \\
& Original & \cellcolor{yellow!25}\textbf{90.55} & \cellcolor{yellow!25}\textbf{70.34} & \cellcolor{yellow!25}\textbf{24.92} & \cellcolor{yellow!25}14.31 \\
\hline
\multirow{5}{*}{\makecell[c]{ResNet-18\\(FMNIST)\\$Acc_O$: 93.63\%}} & Full GC-Net & 91.17 & 24.11 & 45.19 & 23.7 \\
& GC-Net-FH & 91.2 & 21.16 & 34.72 & 19.6 \\
& GC-Net-BH & 92.09 & 31.8 & 61.22 & \textbf{26.73} \\
& GC-Net-B25\% & 92.08 & 32.85 & 61.77 & 23.66 \\
& Original & \cellcolor{yellow!25}\textbf{92.38} & \cellcolor{yellow!25}\textbf{34.33} & \cellcolor{yellow!25}\textbf{64.15} & \cellcolor{yellow!25}25.25 \\
\hline
\multirow{5}{*}{\makecell[c]{ResNet-18\\(Tiny-IN)\\$Acc_O$: 65.15\%}} & Full GC-Net & 34.72 & 2.95 & 1.31 & 2.24 \\
& GC-Net-FH & 51.27 & 5.71 & 2.51 & 4.16 \\
& GC-Net-BH & 46.92 & 5.46 & 2.04 & 4.2 \\
& GC-Net-B25\% & 53.05 & 6.16 & 2.81 & 5.08 \\
& Original & \cellcolor{yellow!25}\textbf{58.17} & \cellcolor{yellow!25}\textbf{7.64} & \cellcolor{yellow!25}\textbf{3.11} & \cellcolor{yellow!25}\textbf{5.97} \\
\hline
\end{tabular}
\caption{ResNet-18 \textbf{OS-SynFlow} average accuracy (\%) across datasets and hybrid types at \textbf{80\% sparsity} using Pearson correlation. Results for applying the pruning method directly to the original model, without a hybrid method, are highlighted in yellow.}
\label{tab:performance_comparison_resnet18_OS_SynFlow_80}
\end{table}

\begin{table}[!ht]
\centering
\setlength{\tabcolsep}{4pt}
\footnotesize
\begin{tabular}{|>{\centering\arraybackslash}p{1.8cm}|c|cccc|}
\hline
\multirow{2}{*}{\makecell[c]{Architecture}} & \multirow{2}{*}{Hybrid Type} & \multicolumn{4}{c|}{\textbf{C-SNIP - Sparsity 20\%}} \\
 &  & Orig($Acc_1$) & CJG & RNB & LO \\
\hline
\multirow{5}{*}{\makecell[c]{ResNet-18\\(CIFAR-10)\\$Acc_O$: 93.14\%}} & Full GC-Net & 93.1 & 73.73 & 24.52 & \textbf{13.82} \\
& GC-Net-FH & 92.48 & 72.13 & 23.35 & 13.61 \\
& GC-Net-BH & \textbf{93.14} & \textbf{73.82} & \textbf{24.76} & 13.74 \\
& GC-Net-B25\% & \textbf{93.14} & 73.75 & 24.51 & 13.74 \\
& Original & \cellcolor{yellow!25}93.08 & \cellcolor{yellow!25}73.77 & \cellcolor{yellow!25}24.62 & \cellcolor{yellow!25}13.64 \\
\hline
\multirow{5}{*}{\makecell[c]{ResNet-18\\(FMNIST)\\$Acc_O$: 93.63\%}} & Full GC-Net & 93.35 & 27.67 & \textbf{54.02} & \textbf{23.31} \\
& GC-Net-FH & 93.17 & \textbf{27.86} & 48.96 & 22.01 \\
& GC-Net-BH & 93.24 & 27.74 & 54.84 & 22.98 \\
& GC-Net-B25\% & 93.24 & 25.73 & 51.46 & 22.83 \\
& Original & \cellcolor{yellow!25}\textbf{93.4} & \cellcolor{yellow!25}25.8 & \cellcolor{yellow!25}47.41 & \cellcolor{yellow!25}19.66 \\
\hline
\multirow{5}{*}{\makecell[c]{ResNet-18\\(Tiny-IN)\\$Acc_O$: 65.15\%}} & Full GC-Net & 62.63 & \textbf{7.08} & 5.08 & \textbf{5.73} \\
& GC-Net-FH & 64.03 & 6.95 & \textbf{5.54} & 5.62 \\
& GC-Net-BH & 62.37 & 7.05 & 4.92 & 5.68 \\
& GC-Net-B25\% & 62.79 & 6.86 & 4.57 & 5.38 \\
& Original & \cellcolor{yellow!25}\textbf{64.39} & \cellcolor{yellow!25}6.91 & \cellcolor{yellow!25}4.93 & \cellcolor{yellow!25}5.58 \\
\hline
\end{tabular}
\caption{ResNet-18 \textbf{C-SNIP} average accuracy (\%) across datasets and hybrid types at \textbf{20\% sparsity}. Results for applying the pruning method directly to the original model, without a hybrid method, are highlighted in yellow.}
\label{tab:performance_comparison_resnet18_C_SNIP_20}
\end{table}

\begin{table}[!ht]
\centering
\setlength{\tabcolsep}{4pt}
\footnotesize
\begin{tabular}{|>{\centering\arraybackslash}p{1.8cm}|c|cccc|}
\hline
\multirow{2}{*}{\makecell[c]{Architecture}} & \multirow{2}{*}{Hybrid Type} & \multicolumn{4}{c|}{\textbf{C-SNIP - Sparsity 40\%}} \\
 &  & Orig($Acc_1$) & CJG & RNB & LO \\
\hline
\multirow{5}{*}{\makecell[c]{ResNet-18\\(CIFAR-10)\\$Acc_O$: 93.14\%}} & Full GC-Net & 93.12 & 73.69 & 24.27 & 13.82 \\
& GC-Net-FH & 91.64 & 70.9 & 23.02 & \textbf{14.14} \\
& GC-Net-BH & 93.15 & 73.69 & \textbf{24.74} & 13.98 \\
& GC-Net-B25\% & \textbf{93.19} & \textbf{73.88} & 24.49 & 13.38 \\
& Original & \cellcolor{yellow!25}93.13 & \cellcolor{yellow!25}73.8 & \cellcolor{yellow!25}24.6 & \cellcolor{yellow!25}13.42 \\
\hline
\multirow{5}{*}{\makecell[c]{ResNet-18\\(FMNIST)\\$Acc_O$: 93.63\%}} & Full GC-Net & 93.14 & 26.09 & 52.84 & 23.1 \\
& GC-Net-FH & 93.09 & \textbf{27.1} & 48.98 & 24.03 \\
& GC-Net-BH & 93.12 & 26.69 & 55.01 & 23.92 \\
& GC-Net-B25\% & 93.3 & 27.08 & \textbf{60.07} & \textbf{24.95} \\
& Original & \cellcolor{yellow!25}\textbf{93.38} & \cellcolor{yellow!25}25.65 & \cellcolor{yellow!25}53.76 & \cellcolor{yellow!25}21.82 \\
\hline
\multirow{5}{*}{\makecell[c]{ResNet-18\\(Tiny-IN)\\$Acc_O$: 65.15\%}} & Full GC-Net & 60.67 & 6.61 & 5.05 & 5.34 \\
& GC-Net-FH & 62.64 & 6.49 & 4.67 & 5.69 \\
& GC-Net-BH & 61.08 & 6.46 & 4.8 & 5.74 \\
& GC-Net-B25\% & 61.91 & 6.82 & 4.61 & 5.5 \\
& Original & \cellcolor{yellow!25}\textbf{64.34} & \cellcolor{yellow!25}\textbf{7.59} & \cellcolor{yellow!25}\textbf{5.09} & \cellcolor{yellow!25}\textbf{5.98} \\
\hline
\end{tabular}
\caption{ResNet-18 \textbf{C-SNIP} average accuracy (\%) across datasets and hybrid types at \textbf{40\% sparsity}. Results for applying the pruning method directly to the original model, without a hybrid method, are highlighted in yellow.}
\label{tab:performance_comparison_resnet18_C_SNIP_40}
\end{table}

\begin{table}[!ht]
\centering
\setlength{\tabcolsep}{4pt}
\footnotesize
\begin{tabular}{|>{\centering\arraybackslash}p{1.8cm}|c|cccc|}
\hline
\multirow{2}{*}{\makecell[c]{Architecture}} & \multirow{2}{*}{Hybrid Type} & \multicolumn{4}{c|}{\textbf{C-SNIP - Sparsity 60\%}} \\
 &  & Orig($Acc_1$) & CJG & RNB & LO \\
\hline
\multirow{5}{*}{\makecell[c]{ResNet-18\\(CIFAR-10)\\$Acc_O$: 93.14\%}} & Full GC-Net & 93.05 & 73.14 & 24.67 & 13.79 \\
& GC-Net-FH & 90.69 & 68.8 & 22.08 & 13.66 \\
& GC-Net-BH & 93.12 & 74.03 & 24.3 & \textbf{13.99} \\
& GC-Net-B25\% & 93.13 & \textbf{74.14} & \textbf{24.95} & 13.8 \\
& Original & \cellcolor{yellow!25}\textbf{93.21} & \cellcolor{yellow!25}73.74 & \cellcolor{yellow!25}24.53 & \cellcolor{yellow!25}13.93 \\
\hline
\multirow{5}{*}{\makecell[c]{ResNet-18\\(FMNIST)\\$Acc_O$: 93.63\%}} & Full GC-Net & 92.87 & \textbf{29.19} & 58.32 & 24.88 \\
& GC-Net-FH & 92.62 & 26.47 & 50.97 & 23.41 \\
& GC-Net-BH & 92.96 & 28.41 & \textbf{58.46} & \textbf{24.9} \\
& GC-Net-B25\% & 93.03 & 26.34 & 56.08 & 24.22 \\
& Original & \cellcolor{yellow!25}\textbf{93.33} & \cellcolor{yellow!25}28.01 & \cellcolor{yellow!25}56.69 & \cellcolor{yellow!25}23.16 \\
\hline
\multirow{5}{*}{\makecell[c]{ResNet-18\\(Tiny-IN)\\$Acc_O$: 65.15\%}} & Full GC-Net & 56.94 & 6.1 & 4.49 & 5.51 \\
& GC-Net-FH & 60.44 & 6.16 & 4.23 & 5.17 \\
& GC-Net-BH & 58.25 & 6.78 & 4.76 & 5.56 \\
& GC-Net-B25\% & 60.32 & 6.82 & 4.66 & 5.58 \\
& Original & \cellcolor{yellow!25}\textbf{63.85} & \cellcolor{yellow!25}\textbf{7.3} & \cellcolor{yellow!25}\textbf{4.85} & \cellcolor{yellow!25}\textbf{5.74} \\
\hline
\end{tabular}
\caption{ResNet-18 \textbf{C-SNIP} average accuracy (\%) across datasets and hybrid types at \textbf{60\% sparsity}. Results for applying the pruning method directly to the original model, without a hybrid method, are highlighted in yellow.}
\label{tab:performance_comparison_resnet18_C_SNIP_60}
\end{table}

\begin{table}[!ht]
\centering
\setlength{\tabcolsep}{4pt}
\footnotesize
\begin{tabular}{|>{\centering\arraybackslash}p{1.8cm}|c|cccc|}
\hline
\multirow{2}{*}{\makecell[c]{Architecture}} & \multirow{2}{*}{Hybrid Type} & \multicolumn{4}{c|}{\textbf{C-SNIP - Sparsity 80\%}} \\
 &  & Orig($Acc_1$) & CJG & RNB & LO \\
\hline
\multirow{5}{*}{\makecell[c]{ResNet-18\\(CIFAR-10)\\$Acc_O$: 93.14\%}} & Full GC-Net & 92.79 & 73.32 & 23.06 & 13.89 \\
& GC-Net-FH & 86.54 & 64.97 & 22.89 & \textbf{14.8} \\
& GC-Net-BH & 92.99 & \textbf{73.84} & 24.97 & 13.94 \\
& GC-Net-B25\% & 93.04 & 73.78 & \textbf{25.85} & 13.94 \\
& Original & \cellcolor{yellow!25}\textbf{93.1} & \cellcolor{yellow!25}73.66 & \cellcolor{yellow!25}25.74 & \cellcolor{yellow!25}14.03 \\
\hline
\multirow{5}{*}{\makecell[c]{ResNet-18\\(FMNIST)\\$Acc_O$: 93.63\%}} & Full GC-Net & 92.84 & 27.15 & 57.15 & 21.58 \\
& GC-Net-FH & 91.85 & 22.49 & 38.99 & 20.08 \\
& GC-Net-BH & 92.82 & \textbf{28.93} & 57.27 & 23.73 \\
& GC-Net-B25\% & 93.01 & 27.76 & 55.25 & 23.62 \\
& Original & \cellcolor{yellow!25}\textbf{93.08} & \cellcolor{yellow!25}27.92 & \cellcolor{yellow!25}\textbf{59.07} & \cellcolor{yellow!25}\textbf{24.19} \\
\hline
\multirow{5}{*}{\makecell[c]{ResNet-18\\(Tiny-IN)\\$Acc_O$: 65.15\%}} & Full GC-Net & 50.64 & 6.06 & 3.45 & 4.74 \\
& GC-Net-FH & 53.11 & 5.68 & 3.31 & 4.47 \\
& GC-Net-BH & 52.77 & 6.43 & 3.35 & 4.75 \\
& GC-Net-B25\% & 56.37 & 6.2 & 3.84 & 5.26 \\
& Original & \cellcolor{yellow!25}\textbf{61.43} & \cellcolor{yellow!25}\textbf{6.99} & \cellcolor{yellow!25}\textbf{4.83} & \cellcolor{yellow!25}\textbf{5.61} \\
\hline
\end{tabular}
\caption{ResNet-18 \textbf{C-SNIP} average accuracy (\%) across datasets and hybrid types at \textbf{80\% sparsity}. Results for applying the pruning method directly to the original model, without a hybrid method, are highlighted in yellow.}
\label{tab:performance_comparison_resnet18_C_SNIP_80}
\end{table}

\begin{table}[!ht]
\centering
\setlength{\tabcolsep}{4pt}
\footnotesize
\begin{tabular}{|>{\centering\arraybackslash}p{1.8cm}|c|cccc|}
\hline
\multirow{2}{*}{\makecell[c]{Architecture}} & \multirow{2}{*}{Hybrid Type} & \multicolumn{4}{c|}{\textbf{L1-Norm - Sparsity 20\%}} \\
 &  & Orig($Acc_1$) & CJG & RNB & LO \\
\hline
\multirow{5}{*}{\makecell[c]{VGG16-BN\\(CIFAR-10)\\ $Acc_O$: 93.18\%}} & Full GC-Net & 91.98 & 73.38 & 27.96 & 13.21 \\
& GC-Net-FH & 92.11 & 73.26 & 28.04 & 12.97 \\
& GC-Net-BH & 93.2 & \textbf{76.53} & 28.93 & 13.13 \\
& GC-Net-B25\% & 93.13 & 76.45 & 29.72 & 13.32 \\
& Original & \cellcolor{yellow!25}\textbf{93.23} & \cellcolor{yellow!25}76.31 & \cellcolor{yellow!25}\textbf{29.8} & \cellcolor{yellow!25}\textbf{13.34} \\
\hline
\multirow{5}{*}{\makecell[c]{VGG16-BN\\(FMNIST)\\ $Acc_O$: 94.36\%}} & Full GC-Net & 93.83 & 34.61 & 65.48 & 25.07 \\
& GC-Net-FH & 93.95 & \textbf{35.46} & 65.77 & 25.34 \\
& GC-Net-BH & \textbf{94.33} & 34.07 & 63.67 & 26.12 \\
& GC-Net-B25\% & 94.2 & 33.08 & \textbf{65.86} & \textbf{26.59} \\
& Original & \cellcolor{yellow!25}94.19 & \cellcolor{yellow!25}33.68 & \cellcolor{yellow!25}65.64 & \cellcolor{yellow!25}26.04 \\
\hline
\multirow{5}{*}{\makecell[c]{VGG16-BN\\(Tiny-IN)\\ $Acc_O$: 64.40\%}} & Full GC-Net & 61.27 & 6.64 & 5.12 & \textbf{5.37} \\
& GC-Net-FH & 62.44 & \textbf{6.86} & 5.12 & 5.06 \\
& GC-Net-BH & 62.03 & 6.15 & 5.87 & 4.78 \\
& GC-Net-B25\% & 62.52 & 6.72 & 5.29 & 5.12 \\
& Original & \cellcolor{yellow!25}\textbf{63.25} & \cellcolor{yellow!25}6.51 & \cellcolor{yellow!25}\textbf{6.97} & \cellcolor{yellow!25}4.8 \\
\hline
\end{tabular}
\caption{VGG16-BN \textbf{L1-Pruning} average accuracy (\%) across datasets and hybrid types at \textbf{20\% sparsity} using Pearson correlation GC-Net. Results for applying the pruning method directly to the original model, without a hybrid method, are highlighted in yellow.}
\label{tab:performance_comparison_vgg16bn_L1_20}
\end{table}

\begin{table}[!ht]
\centering
\setlength{\tabcolsep}{4pt}
\footnotesize
\begin{tabular}{|>{\centering\arraybackslash}p{1.8cm}|c|cccc|}
\hline
\multirow{2}{*}{\makecell[c]{Architecture}} & \multirow{2}{*}{Hybrid Type} & \multicolumn{4}{c|}{\textbf{L1-Norm - Sparsity 40\%}} \\
 &  & Orig($Acc_1$) & CJG & RNB & LO \\
\hline
\multirow{5}{*}{\makecell[c]{VGG16-BN\\(CIFAR-10)\\ $Acc_O$: 93.18\%}} & Full GC-Net & 90.53 & 70.48 & 25.32 & \textbf{13.68} \\
& GC-Net-FH & 90.49 & 70.45 & 24.13 & 13.64 \\
& GC-Net-BH & 92.91 & 75.88 & 29.14 & 13.06 \\
& GC-Net-B25\% & \textbf{93.16} & 76.24 & \textbf{30.15} & 13.24 \\
& Original & \cellcolor{yellow!25}93.15 & \cellcolor{yellow!25}\textbf{76.32} & \cellcolor{yellow!25}29.26 & \cellcolor{yellow!25}10.53 \\
\hline
\multirow{5}{*}{\makecell[c]{VGG16-BN\\(FMNIST)\\ $Acc_O$: 94.36\%}} & Full GC-Net & 92.96 & 32.62 & 61.73 & 23.4 \\
& GC-Net-FH & 93.21 & 33.78 & 61.2 & 23.48 \\
& GC-Net-BH & \textbf{94.21} & 33.5 & 62.0 & 25.5 \\
& GC-Net-B25\% & 94.14 & 34.2 & \textbf{66.08} & \textbf{25.54} \\
& Original & \cellcolor{yellow!25}94.13 & \cellcolor{yellow!25}\textbf{34.34} & \cellcolor{yellow!25}64.85 & \cellcolor{yellow!25}25.37 \\
\hline
\multirow{5}{*}{\makecell[c]{VGG16-BN\\(Tiny-IN)\\ $Acc_O$: 64.40\%}} & Full GC-Net & 56.18 & 5.58 & 5.47 & 4.33 \\
& GC-Net-FH & 60.3 & 6.62 & 5.18 & \textbf{4.87} \\
& GC-Net-BH & 60.14 & 6.2 & 6.24 & 4.13 \\
& GC-Net-B25\% & 62.23 & \textbf{6.64} & 5.36 & 4.73 \\
& Original & \cellcolor{yellow!25}\textbf{63.47} & \cellcolor{yellow!25}6.55 & \cellcolor{yellow!25}\textbf{7.19} & \cellcolor{yellow!25}4.84 \\
\hline
\end{tabular}
\caption{VGG16-BN \textbf{L1-Pruning} average accuracy (\%) across datasets and hybrid types at \textbf{40\% sparsity} using Pearson correlation GC-Net. Results for applying the pruning method directly to the original model, without a hybrid method, are highlighted in yellow.}
\label{tab:performance_comparison_vgg16bn_L1_40}
\end{table}

\begin{table}[!ht]
\centering
\setlength{\tabcolsep}{4pt}
\footnotesize
\begin{tabular}{|>{\centering\arraybackslash}p{1.8cm}|c|cccc|}
\hline
\multirow{2}{*}{\makecell[c]{Architecture}} & \multirow{2}{*}{Hybrid Type} & \multicolumn{4}{c|}{\textbf{L1-Norm - Sparsity 60\%}} \\
 &  & Orig($Acc_1$) & CJG & RNB & LO \\
\hline
\multirow{5}{*}{\makecell[c]{VGG16-BN\\(CIFAR-10)\\ $Acc_O$: 93.18\%}} & Full GC-Net & 87.3 & 65.06 & 23.6 & 12.71 \\
& GC-Net-FH & 87.72 & 65.73 & 21.04 & 12.68 \\
& GC-Net-BH & 92.75 & 75.48 & \textbf{29.23} & 12.49 \\
& GC-Net-B25\% & \textbf{93.19} & \textbf{76.19} & \textbf{29.23} & 12.75 \\
& Original & \cellcolor{yellow!25}93.17 & \cellcolor{yellow!25}75.89 & \cellcolor{yellow!25}28.39 & \cellcolor{yellow!25}\textbf{13.09} \\
\hline
\multirow{5}{*}{\makecell[c]{VGG16-BN\\(FMNIST)\\ $Acc_O$: 94.36\%}} & Full GC-Net & 92.13 & 18.51 & 22.99 & 15.06 \\
& GC-Net-FH & 92.33 & 24.63 & 38.79 & 18.91 \\
& GC-Net-BH & 93.64 & 31.94 & 63.32 & \textbf{25.03} \\
& GC-Net-B25\% & \textbf{94.09} & 33.57 & 65.19 & 24.4 \\
& Original & \cellcolor{yellow!25}93.99 & \cellcolor{yellow!25}\textbf{34.7} & \cellcolor{yellow!25}\textbf{66.19} & \cellcolor{yellow!25}23.88 \\
\hline
\multirow{5}{*}{\makecell[c]{VGG16-BN\\(Tiny-IN)\\ $Acc_O$: 64.40\%}} & Full GC-Net & 46.21 & 5.18 & 2.63 & 3.93 \\
& GC-Net-FH & 55.51 & 6.47 & 3.62 & \textbf{4.98} \\
& GC-Net-BH & 57.4 & 5.64 & 5.59 & 4.22 \\
& GC-Net-B25\% & 61.25 & \textbf{6.59} & 5.31 & 4.71 \\
& Original & \cellcolor{yellow!25}\textbf{63.33} & \cellcolor{yellow!25}6.52 & \cellcolor{yellow!25}\textbf{6.86} & \cellcolor{yellow!25}4.58 \\
\hline
\end{tabular}
\caption{VGG16-BN \textbf{L1-Pruning} average accuracy (\%) across datasets and hybrid types at \textbf{60\% sparsity} using Pearson correlation GC-Net. Results for applying the pruning method directly to the original model, without a hybrid method, are highlighted in yellow.}
\label{tab:performance_comparison_vgg16bn_L1_60}
\end{table}

\begin{table}[!ht]
\centering
\setlength{\tabcolsep}{4pt}
\footnotesize
\begin{tabular}{|>{\centering\arraybackslash}p{1.8cm}|c|cccc|}
\hline
\multirow{2}{*}{\makecell[c]{Architecture}} & \multirow{2}{*}{Hybrid Type} & \multicolumn{4}{c|}{\textbf{L1-Norm - Sparsity 80\%}} \\
 &  & Orig($Acc_1$) & CJG & RNB & LO \\
\hline
\multirow{5}{*}{\makecell[c]{VGG16-BN\\(CIFAR-10)\\ $Acc_O$: 93.18\%}} & Full GC-Net & 58.24 & 42.97 & 14.37 & 10.94 \\
& GC-Net-FH & 69.34 & 52.01 & 16.91 & \textbf{12.32} \\
& GC-Net-BH & 91.81 & 73.07 & 23.6 & 11.79 \\
& GC-Net-B25\% & 92.27 & 74.28 & \textbf{26.79} & 11.89 \\
& Original & \cellcolor{yellow!25}\textbf{92.38} & \cellcolor{yellow!25}\textbf{74.57} & \cellcolor{yellow!25}25.42 & \cellcolor{yellow!25}12.01 \\
\hline
\multirow{5}{*}{\makecell[c]{VGG16-BN\\(FMNIST)\\ $Acc_O$: 94.36\%}} & Full GC-Net & 90.7 & 10.22 & 10.21 & 10.43 \\
& GC-Net-FH & 90.84 & 10.2 & 10.13 & 10.52 \\
& GC-Net-BH & \textbf{93.32} & \textbf{35.24} & \textbf{65.01} & \textbf{23.1} \\
& GC-Net-B25\% & 93.13 & 32.1 & 64.51 & 21.27 \\
& Original & \cellcolor{yellow!25}93.31 & \cellcolor{yellow!25}34.34 & \cellcolor{yellow!25}64.41 & \cellcolor{yellow!25}20.22 \\
\hline
\multirow{5}{*}{\makecell[c]{VGG16-BN\\(Tiny-IN)\\ $Acc_O$: 64.40\%}} & Full GC-Net & 22.31 & 1.86 & 1.26 & 1.68 \\
& GC-Net-FH & 40.64 & 3.51 & 1.45 & 2.87 \\
& GC-Net-BH & 50.14 & 5.58 & 3.75 & 3.72 \\
& GC-Net-B25\% & 56.99 & 6.4 & 4.9 & 4.73 \\
& Original & \cellcolor{yellow!25}\textbf{61.25} & \cellcolor{yellow!25}\textbf{6.64} & \cellcolor{yellow!25}\textbf{7.25} & \cellcolor{yellow!25}\textbf{5.02} \\
\hline
\end{tabular}
\caption{VGG16-BN \textbf{L1-Pruning} average accuracy (\%) across datasets and hybrid types at \textbf{80\% sparsity} using Pearson correlation GC-Net. Results for applying the pruning method directly to the original model, without a hybrid method, are highlighted in yellow.}
\label{tab:performance_comparison_vgg16bn_L1_80}
\end{table}

\begin{table}[!ht]
\centering
\setlength{\tabcolsep}{4pt}
\footnotesize
\begin{tabular}{|>{\centering\arraybackslash}p{1.8cm}|c|cccc|}
\hline
\multirow{2}{*}{\makecell[c]{Architecture}} & \multirow{2}{*}{Hybrid Type} & \multicolumn{4}{c|}{\textbf{L2-Norm - Sparsity 20\%}} \\
 &  & Orig($Acc_1$) & CJG & RNB & LO \\
\hline
\multirow{5}{*}{\makecell[c]{VGG16-BN\\(CIFAR-10)\\ $Acc_O$: 93.18\%}} & Full GC-Net & 90.38 & 69.96 & 26.23 & 14.07 \\
& GC-Net-FH & 90.33 & 69.9 & 25.6 & \textbf{14.16} \\
& GC-Net-BH & 92.1 & 74.09 & \textbf{29.51} & 12.8 \\
& GC-Net-B25\% & \textbf{92.32} & \textbf{74.41} & 29.14 & 12.78 \\
& Original & \cellcolor{yellow!25}92.24 & \cellcolor{yellow!25}73.92 & \cellcolor{yellow!25}28.6 & \cellcolor{yellow!25}12.67 \\
\hline
\multirow{5}{*}{\makecell[c]{VGG16-BN\\(FMNIST)\\ $Acc_O$: 94.36\%}} & Full GC-Net & 93.75 & 29.69 & 59.06 & 21.71 \\
& GC-Net-FH & 93.88 & 30.52 & 60.56 & 23.14 \\
& GC-Net-BH & 94.04 & 35.11 & 64.16 & 27.4 \\
& GC-Net-B25\% & \textbf{94.1} & 36.71 & 63.31 & \textbf{28.53} \\
& Original & \cellcolor{yellow!25}94.04 & \cellcolor{yellow!25}\textbf{37.45} & \cellcolor{yellow!25}\textbf{64.86} & \cellcolor{yellow!25}28.11 \\
\hline
\multirow{5}{*}{\makecell[c]{VGG16-BN\\(Tiny-IN)\\ $Acc_O$: 64.40\%}} & Full GC-Net & 55.97 & 5.98 & 4.96 & 4.52 \\
& GC-Net-FH & 57.33 & 5.57 & 4.64 & 4.67 \\
& GC-Net-BH & 57.68 & 6.8 & 4.91 & \textbf{5.37} \\
& GC-Net-B25\% & 58.69 & \textbf{7.34} & 5.4 & 5.08 \\
& Original & \cellcolor{yellow!25}\textbf{59.1} & \cellcolor{yellow!25}7.08 & \cellcolor{yellow!25}\textbf{7.88} & \cellcolor{yellow!25}4.76 \\
\hline
\end{tabular}
\caption{VGG16-BN \textbf{L2-Pruning} average accuracy (\%) across datasets and hybrid types at \textbf{20\% sparsity} using Pearson correlation GC-Net. Results for applying the pruning method directly to the original model, without a hybrid method, are highlighted in yellow.}
\label{tab:performance_comparison_vgg16bn_L2_20}
\end{table}

\begin{table}[!ht]
\centering
\setlength{\tabcolsep}{4pt}
\footnotesize
\begin{tabular}{|>{\centering\arraybackslash}p{1.8cm}|c|cccc|}
\hline
\multirow{2}{*}{\makecell[c]{Architecture}} & \multirow{2}{*}{Hybrid Type} & \multicolumn{4}{c|}{\textbf{L2-Norm - Sparsity 40\%}} \\
 &  & Orig($Acc_1$) & CJG & RNB & LO \\
\hline
\multirow{5}{*}{\makecell[c]{VGG16-BN\\(CIFAR-10)\\ $Acc_O$: 93.18\%}} & Full GC-Net & 87.15 & 65.6 & 26.73 & 14.25 \\
& GC-Net-FH & 87.35 & 66.54 & 26.12 & \textbf{14.58} \\
& GC-Net-BH & 90.66 & \textbf{70.31} & 27.38 & 13.27 \\
& GC-Net-B25\% & \textbf{90.88} & 69.84 & 28.0 & 13.11 \\
& Original & \cellcolor{yellow!25}90.8 & \cellcolor{yellow!25}69.87 & \cellcolor{yellow!25}\textbf{28.21} & \cellcolor{yellow!25}12.91 \\
\hline
\multirow{5}{*}{\makecell[c]{VGG16-BN\\(FMNIST)\\ $Acc_O$: 94.36\%}} & Full GC-Net & 92.9 & 29.85 & 50.94 & 21.34 \\
& GC-Net-FH & 93.01 & 27.92 & 56.48 & 20.4 \\
& GC-Net-BH & 93.57 & \textbf{33.82} & 59.66 & 30.57 \\
& GC-Net-B25\% & 93.61 & 33.69 & 55.16 & \textbf{42.8} \\
& Original & \cellcolor{yellow!25}\textbf{93.77} & \cellcolor{yellow!25}33.5 & \cellcolor{yellow!25}\textbf{66.28} & \cellcolor{yellow!25}31.06 \\
\hline
\multirow{5}{*}{\makecell[c]{VGG16-BN\\(Tiny-IN)\\ $Acc_O$: 64.40\%}} & Full GC-Net & 43.13 & 4.34 & 2.57 & 3.51 \\
& GC-Net-FH & 45.33 & 5.25 & 2.69 & 3.97 \\
& GC-Net-BH & 48.67 & 6.93 & 3.19 & 5.3 \\
& GC-Net-B25\% & 49.79 & 7.03 & 3.36 & \textbf{5.72} \\
& Original & \cellcolor{yellow!25}\textbf{51.05} & \cellcolor{yellow!25}\textbf{7.08} & \cellcolor{yellow!25}\textbf{6.61} & \cellcolor{yellow!25}5.66 \\
\hline
\end{tabular}
\caption{VGG16-BN \textbf{L2-Pruning} average accuracy (\%) across datasets and hybrid types at \textbf{40\% sparsity} using Pearson correlation GC-Net. Results for applying the pruning method directly to the original model, without a hybrid method, are highlighted in yellow.}
\label{tab:performance_comparison_vgg16bn_L2_40}
\end{table}

\begin{table}[!ht]
\centering
\setlength{\tabcolsep}{4pt}
\footnotesize
\begin{tabular}{|>{\centering\arraybackslash}p{1.8cm}|c|cccc|}
\hline
\multirow{2}{*}{\makecell[c]{Architecture}} & \multirow{2}{*}{Hybrid Type} & \multicolumn{4}{c|}{\textbf{L2-Norm - Sparsity 60\%}} \\
 &  & Orig($Acc_1$) & CJG & RNB & LO \\
\hline
\multirow{5}{*}{\makecell[c]{VGG16-BN\\(CIFAR-10)\\ $Acc_O$: 93.18\%}} & Full GC-Net & 80.83 & 58.31 & 23.37 & 14.31 \\
& GC-Net-FH & 82.56 & 60.05 & 22.92 & \textbf{14.89} \\
& GC-Net-BH & 85.91 & 61.61 & 24.07 & 14.62 \\
& GC-Net-B25\% & 86.09 & 61.95 & 23.98 & 14.83 \\
& Original & \cellcolor{yellow!25}\textbf{86.68} & \cellcolor{yellow!25}\textbf{62.14} & \cellcolor{yellow!25}\textbf{24.92} & \cellcolor{yellow!25}14.72 \\
\hline
\multirow{5}{*}{\makecell[c]{VGG16-BN\\(FMNIST)\\ $Acc_O$: 94.36\%}} & Full GC-Net & 91.88 & 27.46 & 51.95 & 20.4 \\
& GC-Net-FH & 92.18 & 25.32 & 36.83 & 19.18 \\
& GC-Net-BH & 92.55 & 32.69 & 64.9 & \textbf{33.47} \\
& GC-Net-B25\% & \textbf{92.91} & \textbf{33.62} & \textbf{66.38} & 33.01 \\
& Original & \cellcolor{yellow!25}92.59 & \cellcolor{yellow!25}33.18 & \cellcolor{yellow!25}65.85 & \cellcolor{yellow!25}31.97 \\
\hline
\multirow{5}{*}{\makecell[c]{VGG16-BN\\(Tiny-IN)\\ $Acc_O$: 64.40\%}} & Full GC-Net & 23.31 & 2.69 & 1.32 & 2.0 \\
& GC-Net-FH & 22.33 & 2.44 & 1.45 & 1.64 \\
& GC-Net-BH & 28.01 & 4.91 & 1.49 & 3.29 \\
& GC-Net-B25\% & 29.15 & 4.72 & 1.46 & 3.32 \\
& Original & \cellcolor{yellow!25}\textbf{29.47} & \cellcolor{yellow!25}\textbf{5.09} & \cellcolor{yellow!25}\textbf{3.93} & \cellcolor{yellow!25}\textbf{3.88} \\
\hline
\end{tabular}
\caption{VGG16-BN \textbf{L2-Pruning} average accuracy (\%) across datasets and hybrid types at \textbf{60\% sparsity} using Pearson correlation GC-Net. Results for applying the pruning method directly to the original model, without a hybrid method, are highlighted in yellow.}
\label{tab:performance_comparison_vgg16bn_L2_60}
\end{table}

\begin{table}[!ht]
\centering
\setlength{\tabcolsep}{4pt}
\footnotesize
\begin{tabular}{|>{\centering\arraybackslash}p{1.8cm}|c|cccc|}
\hline
\multirow{2}{*}{\makecell[c]{Architecture}} & \multirow{2}{*}{Hybrid Type} & \multicolumn{4}{c|}{\textbf{L2-Norm - Sparsity 80\%}} \\
 &  & Orig($Acc_1$) & CJG & RNB & LO \\
\hline
\multirow{5}{*}{\makecell[c]{VGG16-BN\\(CIFAR-10)\\ $Acc_O$: 93.18\%}} & Full GC-Net & 48.5 & 33.26 & 14.76 & 12.64 \\
& GC-Net-FH & 67.86 & 44.71 & 19.3 & 13.1 \\
& GC-Net-BH & 45.72 & 30.23 & 15.14 & 14.36 \\
& GC-Net-B25\% & 36.17 & 25.11 & 13.84 & 14.06 \\
& Original & \cellcolor{yellow!25}\textbf{74.88} & \cellcolor{yellow!25}\textbf{46.33} & \cellcolor{yellow!25}\textbf{20.55} & \cellcolor{yellow!25}\textbf{15.18} \\
\hline
\multirow{5}{*}{\makecell[c]{VGG16-BN\\(FMNIST)\\ $Acc_O$: 94.36\%}} & Full GC-Net & 90.89 & 11.72 & 16.38 & 12.95 \\
& GC-Net-FH & 90.61 & 11.92 & 14.77 & 11.46 \\
& GC-Net-BH & \textbf{91.66} & 32.14 & 57.73 & 30.95 \\
& GC-Net-B25\% & 91.53 & \textbf{33.34} & 62.66 & 34.19 \\
& Original & \cellcolor{yellow!25}91.48 & \cellcolor{yellow!25}32.58 & \cellcolor{yellow!25}\textbf{64.1} & \cellcolor{yellow!25}\textbf{34.27} \\
\hline
\multirow{5}{*}{\makecell[c]{VGG16-BN\\(Tiny-IN)\\ $Acc_O$: 64.40\%}} & Full GC-Net & 9.92 & 1.55 & 0.71 & 1.23 \\
& GC-Net-FH & 2.77 & 0.96 & 0.67 & 1.14 \\
& GC-Net-BH & \textbf{5.84} & \textbf{1.85} & \textbf{0.85} & \textbf{1.36} \\
& GC-Net-B25\% & 3.78 & 1.29 & 0.67 & 1.34 \\
& Original & \cellcolor{yellow!25}2.42 & \cellcolor{yellow!25}0.83 & \cellcolor{yellow!25}0.65 & \cellcolor{yellow!25}0.87 \\
\hline
\end{tabular}
\caption{VGG16-BN \textbf{L2-Pruning} average accuracy (\%) across datasets and hybrid types at \textbf{80\% sparsity} using Pearson correlation GC-Net. Results for applying the pruning method directly to the original model, without a hybrid method, are highlighted in yellow.}
\label{tab:performance_comparison_vgg16bn_L2_80}
\end{table}

\begin{table}[!ht]
\centering
\setlength{\tabcolsep}{4pt}
\footnotesize
\begin{tabular}{|>{\centering\arraybackslash}p{1.8cm}|c|cccc|}
\hline
\multirow{2}{*}{\makecell[c]{Architecture}} & \multirow{2}{*}{Hybrid Type} & \multicolumn{4}{c|}{\textbf{C-SNIP - Sparsity 20\%}} \\
 &  & Orig($Acc_1$) & CJG & RNB & LO \\
\hline
\multirow{5}{*}{\makecell[c]{VGG16-BN\\(CIFAR-10)\\ $Acc_O$: 93.18\%}} & Full GC-Net & 92.71 & 75.85 & 29.96 & 13.32 \\
& GC-Net-FH & 93.01 & \textbf{76.71} & 29.79 & \textbf{13.43} \\
& GC-Net-BH & 93.08 & 76.44 & \textbf{30.16} & 13.39 \\
& GC-Net-B25\% & \textbf{93.32} & 76.61 & 29.95 & 12.86 \\
& Original & \cellcolor{yellow!25}93.23 & \cellcolor{yellow!25}76.41 & \cellcolor{yellow!25}29.62 & \cellcolor{yellow!25}13.18 \\
\hline
\multirow{5}{*}{\makecell[c]{VGG16-BN\\(FMNIST)\\ $Acc_O$: 94.36\%}} & Full GC-Net & 94.14 & 34.51 & 64.5 & 24.23 \\
& GC-Net-FH & \textbf{94.3} & 34.92 & 64.91 & \textbf{26.24} \\
& GC-Net-BH & 94.17 & \textbf{36.28} & 63.43 & 26.17 \\
& GC-Net-B25\% & 94.19 & 33.39 & 64.42 & 26.03 \\
& Original & \cellcolor{yellow!25}94.22 & \cellcolor{yellow!25}34.03 & \cellcolor{yellow!25}\textbf{65.31} & \cellcolor{yellow!25}26.23 \\
\hline
\multirow{5}{*}{\makecell[c]{VGG16-BN\\(Tiny-IN)\\ $Acc_O$: 64.40\%}} & Full GC-Net & 60.79 & 6.11 & \textbf{5.69} & 4.45 \\
& GC-Net-FH & 60.96 & 5.8 & 5.46 & 4.3 \\
& GC-Net-BH & 61.87 & 6.06 & 5.41 & 4.48 \\
& GC-Net-B25\% & \textbf{63.05} & \textbf{6.58} & 5.64 & 4.64 \\
& Original & \cellcolor{yellow!25}63.03 & \cellcolor{yellow!25}6.5 & \cellcolor{yellow!25}5.35 & \cellcolor{yellow!25}\textbf{4.89} \\
\hline
\end{tabular}
\caption{VGG16-BN \textbf{C-SNIP} average accuracy (\%) across datasets and hybrid types at \textbf{20\% sparsity} using Pearson correlation. Results for applying the pruning method directly to the original model, without a hybrid method, are highlighted in yellow.}
\label{tab:performance_comparison_vgg16bn_C_SNIP_pearson_20}
\end{table}

\begin{table}[!ht]
\centering
\setlength{\tabcolsep}{4pt}
\footnotesize
\begin{tabular}{|>{\centering\arraybackslash}p{1.8cm}|c|cccc|}
\hline
\multirow{2}{*}{\makecell[c]{Architecture}} & \multirow{2}{*}{Hybrid Type} & \multicolumn{4}{c|}{\textbf{C-SNIP - Sparsity 40\%}} \\
 &  & Orig($Acc_1$) & CJG & RNB & LO \\
\hline
\multirow{5}{*}{\makecell[c]{VGG16-BN\\(CIFAR-10)\\ $Acc_O$: 93.18\%}} & Full GC-Net & 92.87 & 75.69 & 30.03 & 13.34 \\
& GC-Net-FH & 92.73 & 75.62 & 29.91 & 13.16 \\
& GC-Net-BH & 93.0 & 75.88 & \textbf{30.17} & 12.92 \\
& GC-Net-B25\% & \textbf{93.24} & \textbf{76.37} & 29.77 & 13.2 \\
& Original & \cellcolor{yellow!25}93.2 & \cellcolor{yellow!25}76.21 & \cellcolor{yellow!25}29.47 & \cellcolor{yellow!25}\textbf{13.45} \\
\hline
\multirow{5}{*}{\makecell[c]{VGG16-BN\\(FMNIST)\\ $Acc_O$: 94.36\%}} & Full GC-Net & 94.0 & 33.26 & 62.47 & 23.37 \\
& GC-Net-FH & 94.23 & \textbf{35.42} & 64.07 & 24.76 \\
& GC-Net-BH & \textbf{94.24} & 35.21 & 63.67 & 26.21 \\
& GC-Net-B25\% & 94.17 & 33.49 & 64.76 & 25.88 \\
& Original & \cellcolor{yellow!25}\textbf{94.24} & \cellcolor{yellow!25}34.02 & \cellcolor{yellow!25}\textbf{64.81} & \cellcolor{yellow!25}\textbf{26.51} \\
\hline
\multirow{5}{*}{\makecell[c]{VGG16-BN\\(Tiny-IN)\\ $Acc_O$: 64.40\%}} & Full GC-Net & 60.51 & 5.91 & \textbf{5.86} & 4.24 \\
& GC-Net-FH & 57.75 & 5.07 & 5.17 & 3.07 \\
& GC-Net-BH & 61.42 & 6.37 & 5.61 & 4.59 \\
& GC-Net-B25\% & 62.84 & \textbf{6.72} & 5.6 & \textbf{5.01} \\
& Original & \cellcolor{yellow!25}\textbf{62.89} & \cellcolor{yellow!25}6.46 & \cellcolor{yellow!25}5.4 & \cellcolor{yellow!25}4.68 \\
\hline
\end{tabular}
\caption{VGG16-BN \textbf{C-SNIP} average accuracy (\%) across datasets and hybrid types at \textbf{40\% sparsity} using Pearson correlation. Results for applying the pruning method directly to the original model, without a hybrid method, are highlighted in yellow.}
\label{tab:performance_comparison_vgg16bn_C_SNIP_pearson_40}
\end{table}

\begin{table}[!ht]
\centering
\setlength{\tabcolsep}{4pt}
\footnotesize
\begin{tabular}{|>{\centering\arraybackslash}p{1.8cm}|c|cccc|}
\hline
\multirow{2}{*}{\makecell[c]{Architecture}} & \multirow{2}{*}{Hybrid Type} & \multicolumn{4}{c|}{\textbf{C-SNIP - Sparsity 60\%}} \\
 &  & Orig($Acc_1$) & CJG & RNB & LO \\
\hline
\multirow{5}{*}{\makecell[c]{VGG16-BN\\(CIFAR-10)\\ $Acc_O$: 93.18\%}} & Full GC-Net & 92.6 & 75.31 & 29.75 & 13.46 \\
& GC-Net-FH & 92.19 & 74.3 & 29.2 & \textbf{13.52} \\
& GC-Net-BH & 92.87 & 75.81 & \textbf{30.17} & 12.8 \\
& GC-Net-B25\% & \textbf{93.19} & \textbf{76.37} & 29.47 & 12.58 \\
& Original & \cellcolor{yellow!25}93.18 & \cellcolor{yellow!25}76.0 & \cellcolor{yellow!25}29.53 & \cellcolor{yellow!25}13.36 \\
\hline
\multirow{5}{*}{\makecell[c]{VGG16-BN\\(FMNIST)\\ $Acc_O$: 94.36\%}} & Full GC-Net & 93.94 & 34.76 & 63.27 & 22.87 \\
& GC-Net-FH & 93.98 & 34.07 & 63.47 & 22.57 \\
& GC-Net-BH & 94.26 & \textbf{35.67} & 63.26 & 24.89 \\
& GC-Net-B25\% & 94.08 & 32.97 & 64.79 & 25.09 \\
& Original & \cellcolor{yellow!25}\textbf{94.29} & \cellcolor{yellow!25}33.66 & \cellcolor{yellow!25}\textbf{66.44} & \cellcolor{yellow!25}\textbf{27.51} \\
\hline
\multirow{5}{*}{\makecell[c]{VGG16-BN\\(Tiny-IN)\\ $Acc_O$: 64.40\%}} & Full GC-Net & 58.25 & 5.33 & 5.51 & 3.58 \\
& GC-Net-FH & 52.84 & 4.6 & 4.0 & 3.27 \\
& GC-Net-BH & 60.01 & 6.08 & \textbf{5.98} & 4.2 \\
& GC-Net-B25\% & 62.18 & \textbf{6.63} & 5.57 & \textbf{5.06} \\
& Original & \cellcolor{yellow!25}\textbf{62.92} & \cellcolor{yellow!25}6.54 & \cellcolor{yellow!25}5.7 & \cellcolor{yellow!25}4.68 \\
\hline
\end{tabular}
\caption{VGG16-BN \textbf{C-SNIP} average accuracy (\%) across datasets and hybrid types at \textbf{60\% sparsity} using Pearson correlation. Results for applying the pruning method directly to the original model, without a hybrid method, are highlighted in yellow.}
\label{tab:performance_comparison_vgg16bn_C_SNIP_pearson_60}
\end{table}

\begin{table}[!ht]
\centering
\setlength{\tabcolsep}{4pt}
\footnotesize
\begin{tabular}{|>{\centering\arraybackslash}p{1.8cm}|c|cccc|}
\hline
\multirow{2}{*}{\makecell[c]{Architecture}} & \multirow{2}{*}{Hybrid Type} & \multicolumn{4}{c|}{\textbf{C-SNIP - Sparsity 80\%}} \\
 &  & Orig($Acc_1$) & CJG & RNB & LO \\
\hline
\multirow{5}{*}{\makecell[c]{VGG16-BN\\(CIFAR-10)\\ $Acc_O$: 93.18\%}} & Full GC-Net & 92.3 & 75.11 & 29.6 & 13.58 \\
& GC-Net-FH & 90.9 & 72.43 & 26.66 & \textbf{13.76} \\
& GC-Net-BH & 92.49 & 74.55 & 27.14 & 11.7 \\
& GC-Net-B25\% & 92.75 & 75.14 & 27.51 & 11.91 \\
& Original & \cellcolor{yellow!25}\textbf{93.3} & \cellcolor{yellow!25}\textbf{76.71} & \cellcolor{yellow!25}\textbf{29.95} & \cellcolor{yellow!25}12.93 \\
\hline
\multirow{5}{*}{\makecell[c]{VGG16-BN\\(FMNIST)\\ $Acc_O$: 94.36\%}} & Full GC-Net & 94.0 & \textbf{34.71} & 62.77 & 20.45 \\
& GC-Net-FH & 93.71 & 33.67 & \textbf{67.43} & 23.22 \\
& GC-Net-BH & 94.11 & 34.43 & 64.11 & 23.3 \\
& GC-Net-B25\% & 94.02 & 33.18 & 64.3 & 22.96 \\
& Original & \cellcolor{yellow!25}\textbf{94.25} & \cellcolor{yellow!25}34.46 & \cellcolor{yellow!25}66.35 & \cellcolor{yellow!25}\textbf{28.13} \\
\hline
\multirow{5}{*}{\makecell[c]{VGG16-BN\\(Tiny-IN)\\ $Acc_O$: 64.40\%}} & Full GC-Net & 53.79 & 5.26 & 4.95 & 3.37 \\
& GC-Net-FH & 40.69 & 3.79 & 1.87 & 2.94 \\
& GC-Net-BH & 55.31 & 5.46 & 4.38 & 3.63 \\
& GC-Net-B25\% & 59.12 & 6.47 & 4.94 & 4.79 \\
& Original & \cellcolor{yellow!25}\textbf{63.21} & \cellcolor{yellow!25}\textbf{6.84} & \cellcolor{yellow!25}\textbf{5.95} & \cellcolor{yellow!25}\textbf{5.09} \\
\hline
\end{tabular}%
\caption{VGG16-BN \textbf{OS-SynFlow} average accuracy (\%) across datasets and hybrid types at \textbf{20\% sparsity} using Pearson correlation. Results for applying the pruning method directly to the original model, without a hybrid method, are highlighted in yellow.}
\label{tab:performance_comparison_vgg16bn_OS-SynFlow_20}
\end{table}

\begin{table}[!ht]
\centering
\setlength{\tabcolsep}{4pt}
\footnotesize
\begin{tabular}{|>{\centering\arraybackslash}p{1.8cm}|c|cccc|}
\hline
\multirow{2}{*}{\makecell[c]{Architecture}} & \multirow{2}{*}{Hybrid Type} & \multicolumn{4}{c|}{\textbf{OS-SynFlow - Sparsity 40\%}} \\
 &  & Orig($Acc_1$) & CJG & RNB & LO \\
\hline
\multirow{5}{*}{\makecell[c]{VGG16-BN\\(CIFAR-10)\\ $Acc_O$: 93.18\%}} & Full GC-Net & 89.91 & 69.15 & 26.5 & \textbf{13.54} \\
& GC-Net-FH & 90.17 & 46.38 & 25.57 & 13.0 \\
& GC-Net-BH & 92.98 & \textbf{76.38} & 28.66 & 12.95 \\
& GC-Net-B25\% & 93.27 & 76.29 & 29.42 & 12.92 \\
& Original & \cellcolor{yellow!25}\textbf{93.29} & \cellcolor{yellow!25}76.18 & \cellcolor{yellow!25}\textbf{29.63} & \cellcolor{yellow!25}12.97 \\
\hline
\multirow{5}{*}{\makecell[c]{VGG16-BN\\(FMNIST)\\ $Acc_O$: 94.36\%}} & Full GC-Net & 93.13 & 32.36 & 61.28 & 24.33 \\
& GC-Net-FH & 93.29 & 32.78 & 60.92 & 24.78 \\
& GC-Net-BH & 94.14 & 32.05 & 64.43 & \textbf{26.1} \\
& GC-Net-B25\% & 94.1 & \textbf{34.21} & 67.05 & 25.32 \\
& Original & \cellcolor{yellow!25}\textbf{94.15} & \cellcolor{yellow!25}33.9 & \cellcolor{yellow!25}\textbf{67.21} & \cellcolor{yellow!25}25.14 \\
\hline
\multirow{5}{*}{\makecell[c]{VGG16-BN\\(Tiny-IN)\\ $Acc_O$: 64.40\%}} & Full GC-Net & 56.1 & 6.45 & 3.74 & \textbf{4.94} \\
& GC-Net-FH & 56.58 & 6.15 & 4.12 & 4.35 \\
& GC-Net-BH & 60.97 & 6.39 & \textbf{5.82} & 4.55 \\
& GC-Net-B25\% & 62.64 & \textbf{6.57} & 5.56 & 4.84 \\
& Original & \cellcolor{yellow!25}\textbf{63.39} & \cellcolor{yellow!25}6.25 & \cellcolor{yellow!25}5.55 & \cellcolor{yellow!25}4.64 \\
\hline
\end{tabular}
\caption{VGG16-BN \textbf{OS-SynFlow} average accuracy (\%) across datasets and hybrid types at \textbf{40\% sparsity} using Pearson correlation. Results for applying the pruning method directly to the original model, without a hybrid method, are highlighted in yellow.}
\label{tab:performance_comparison_vgg16bn_OS-SynFlow_40}
\end{table}

\begin{table}[!ht]
\centering
\setlength{\tabcolsep}{4pt}
\footnotesize
%
\caption{VGG16-BN \textbf{OS-SynFlow} average accuracy (\%) across datasets and hybrid types at \textbf{60\% sparsity} using Pearson correlation. Results for applying the pruning method directly to the original model, without a hybrid method, are highlighted in yellow.}
\label{tab:performance_comparison_vgg16bn_OS-SynFlow_60}
\end{table}

\begin{table}[!ht]
\centering
\setlength{\tabcolsep}{4pt}
\footnotesize
%
\caption{VGG16-BN \textbf{OS-SynFlow} average accuracy (\%) across datasets and hybrid types at \textbf{80\% sparsity} using Pearson correlation. Results for applying the pruning method directly to the original model, without a hybrid method, are highlighted in yellow.}
\label{tab:performance_comparison_vgg16bn_OS-SynFlow_80}
\end{table}

\clearpage

\subsection{Additional Experiments - Cosine Similarity}\label{subsec8}
Additional results for OS-SynFlow with cosine similarity for ResNet-18 and C-SNIP with cosine similarity for VGG16-BN.

\begin{table}[!ht]
\centering
\setlength{\tabcolsep}{4pt}
\footnotesize
\begin{tabular}{|>{\centering\arraybackslash}p{1.8cm}|c|cccc|}
\hline
\multirow{2}{*}{\makecell[c]{Architecture}} & \multirow{2}{*}{Hybrid Type} & \multicolumn{4}{c|}{\textbf{OS-SynFlow - Sparsity 20\%}} \\
 &  & Orig($Acc_1$) & CJG & RNB & LO \\
\hline
\multirow{5}{*}{\makecell[c]{ResNet-18\\(CIFAR-10)\\$Acc_O$: 93.14\%}} & Full GC-Net & 92.27 & 72.55 & 25.0 & 13.1 \\
& GC-Net-FH & 92.23 & 71.85 & 23.61 & 13.77 \\
& GC-Net-BH & 93.01 & 73.93 & 25.65 & 13.82 \\
& GC-Net-B25\% & 93.02 & \textbf{73.97} & \textbf{26.26} & \textbf{14.01} \\
& Original & \cellcolor{yellow!25}\textbf{93.15} & \cellcolor{yellow!25}73.87 & \cellcolor{yellow!25}24.48 & \cellcolor{yellow!25}13.32 \\
\hline
\multirow{5}{*}{\makecell[c]{ResNet-18\\(FMNIST)\\$Acc_O$: 93.63\%}} & Full GC-Net & 93.12 & \textbf{27.67} & 57.14 & \textbf{22.24} \\
& GC-Net-FH & 93.18 & 27.0 & \textbf{59.88} & 21.78 \\
& GC-Net-BH & 93.01 & 26.2 & 55.31 & 21.4 \\
& GC-Net-B25\% & \textbf{93.3} & 26.33 & 52.29 & 21.2 \\
& Original & \cellcolor{yellow!25}93.25 & \cellcolor{yellow!25}24.92 & \cellcolor{yellow!25}52.48 & \cellcolor{yellow!25}21.05 \\
\hline
\multirow{5}{*}{\makecell[c]{ResNet-18\\(Tiny-IN)\\$Acc_O$: 65.15\%}} & Full GC-Net & 62.55 & 7.04 & 4.63 & 5.51 \\
& GC-Net-FH & 64.14 & 6.94 & 4.66 & 5.61 \\
& GC-Net-BH & 63.15 & 7.01 & 4.62 & 5.56 \\
& GC-Net-B25\% & 63.62 & \textbf{7.36} & 4.79 & \textbf{5.91} \\
& Original & \cellcolor{yellow!25}\textbf{64.39} & \cellcolor{yellow!25}7.07 & \cellcolor{yellow!25}\textbf{4.8} & \cellcolor{yellow!25}5.59 \\
\hline
\end{tabular}
\caption{ResNet-18 \textbf{OS-SynFlow} average accuracy (\%) across datasets and hybrid types at \textbf{20\% sparsity} using cosine similarity. Results for applying the pruning method directly to the original model, without a hybrid method, are highlighted in yellow.}
\label{tab:performance_comparison_resnet18_OS_SynFlow_cosine_20}
\end{table}

\begin{table}[!ht]
\centering
\setlength{\tabcolsep}{4pt}
\footnotesize
\begin{tabular}{|>{\centering\arraybackslash}p{1.8cm}|c|cccc|}
\hline
\multirow{2}{*}{\makecell[c]{Architecture}} & \multirow{2}{*}{Hybrid Type} & \multicolumn{4}{c|}{\textbf{OS-SynFlow - Sparsity 40\%}} \\
 &  & Orig($Acc_1$) & CJG & RNB & LO \\
\hline
\multirow{5}{*}{\makecell[c]{ResNet-18\\(CIFAR-10)\\$Acc_O$: 93.14\%}} & Full GC-Net & 90.77 & 70.15 & 24.63 & 14.5 \\
& GC-Net-FH & 91.36 & 70.28 & 23.08 & 14.03 \\
& GC-Net-BH & 92.66 & 73.56 & 25.84 & 13.54 \\
& GC-Net-B25\% & \textbf{92.85} & \textbf{74.71} & \textbf{26.27} & \textbf{13.74} \\
& Original & \cellcolor{yellow!25}92.8 & \cellcolor{yellow!25}73.78 & \cellcolor{yellow!25}24.17 & \cellcolor{yellow!25}13.58 \\
\hline
\multirow{5}{*}{\makecell[c]{ResNet-18\\(FMNIST)\\$Acc_O$: 93.63\%}} & Full GC-Net & 92.4 & 26.57 & 54.53 & 24.56 \\
& GC-Net-FH & 92.93 & \textbf{29.05} & 58.06 & \textbf{25.69} \\
& GC-Net-BH & 93.09 & 29.08 & \textbf{59.83} & 23.07 \\
& GC-Net-B25\% & 93.09 & 28.09 & 58.59 & 21.73 \\
& Original & \cellcolor{yellow!25}\textbf{93.43} & \cellcolor{yellow!25}25.54 & \cellcolor{yellow!25}52.69 & \cellcolor{yellow!25}21.36 \\
\hline
\multirow{5}{*}{\makecell[c]{ResNet-18\\(Tiny-IN)\\$Acc_O$: 65.15\%}} & Full GC-Net & 58.62 & 6.94 & 4.14 & 5.59 \\
& GC-Net-FH & 62.9 & 7.15 & 4.51 & 5.48 \\
& GC-Net-BH & 60.87 & 6.62 & 4.42 & 5.21 \\
& GC-Net-B25\% & 62.94 & 7.2 & 4.47 & \textbf{5.73} \\
& Original & \cellcolor{yellow!25}\textbf{64.25} & \cellcolor{yellow!25}\textbf{7.5} & \cellcolor{yellow!25}\textbf{4.5} & \cellcolor{yellow!25}5.49 \\
\hline
\end{tabular}
\caption{ResNet-18 \textbf{OS-SynFlow} average accuracy (\%) across datasets and hybrid types at \textbf{40\% sparsity} using cosine similarity. Results for applying the pruning method directly to the original model, without a hybrid method, are highlighted in yellow.}
\label{tab:performance_comparison_resnet18_OS_SynFlow_cosine_40}
\end{table}

\begin{table}[!ht]
\centering
\setlength{\tabcolsep}{4pt}
\footnotesize
\begin{tabular}{|>{\centering\arraybackslash}p{1.8cm}|c|cccc|}
\hline
\multirow{2}{*}{\makecell[c]{Architecture}} & \multirow{2}{*}{Hybrid Type} & \multicolumn{4}{c|}{\textbf{OS-SynFlow - Sparsity 60\%}} \\
 &  & Orig($Acc_1$) & CJG & RNB & LO \\
\hline
\multirow{5}{*}{\makecell[c]{ResNet-18\\(CIFAR-10)\\$Acc_O$: 93.14\%}} & Full GC-Net & 87.44 & 64.73 & 23.4 & 13.92 \\
& GC-Net-FH & 89.63 & 67.12 & 22.88 & 14.46 \\
& GC-Net-BH & 91.55 & 72.21 & 24.99 & 13.63 \\
& GC-Net-B25\% & 92.19 & 73.16 & \textbf{26.11} & \textbf{14.23} \\
& Original & \cellcolor{yellow!25}\textbf{92.36} & \cellcolor{yellow!25}\textbf{73.17} & \cellcolor{yellow!25}23.01 & \cellcolor{yellow!25}13.54 \\
\hline
\multirow{5}{*}{\makecell[c]{ResNet-18\\(FMNIST)\\$Acc_O$: 93.63\%}} & Full GC-Net & 91.36 & 24.95 & 47.72 & 22.48 \\
& GC-Net-FH & 92.06 & 28.93 & 54.79 & \textbf{26.94} \\
& GC-Net-BH & 92.59 & \textbf{30.59} & \textbf{62.12} & 25.51 \\
& GC-Net-B25\% & 92.71 & 29.61 & 59.45 & 24.01 \\
& Original & \cellcolor{yellow!25}\textbf{92.89} & \cellcolor{yellow!25}25.98 & \cellcolor{yellow!25}52.89 & \cellcolor{yellow!25}21.45 \\
\hline
\multirow{5}{*}{\makecell[c]{ResNet-18\\(Tiny-IN)\\$Acc_O$: 65.15\%}} & Full GC-Net & 49.5 & 5.72 & 2.98 & 4.49 \\
& GC-Net-FH & 59.4 & \textbf{7.48} & 3.64 & 5.72 \\
& GC-Net-BH & 54.87 & 6.57 & 3.49 & 5.19 \\
& GC-Net-B25\% & 58.72 & 6.65 & 3.42 & \textbf{5.92} \\
& Original & \cellcolor{yellow!25}\textbf{63.35} & \cellcolor{yellow!25}7.35 & \cellcolor{yellow!25}\textbf{4.06} & \cellcolor{yellow!25}\textbf{5.92} \\
\hline
\end{tabular}
\caption{ResNet-18 \textbf{OS-SynFlow} average accuracy (\%) across datasets and hybrid types at \textbf{60\% sparsity} using cosine similarity. Results for applying the pruning method directly to the original model, without a hybrid method, are highlighted in yellow.}
\label{tab:performance_comparison_resnet18_OS_SynFlow_cosine_60}
\end{table}

\begin{table}[!ht]
\centering
\setlength{\tabcolsep}{4pt}
\footnotesize
\begin{tabular}{|>{\centering\arraybackslash}p{1.8cm}|c|cccc|}
\hline
\multirow{2}{*}{\makecell[c]{Architecture}} & \multirow{2}{*}{Hybrid Type} & \multicolumn{4}{c|}{\textbf{OS-SynFlow - Sparsity 80\%}} \\
 &  & Orig($Acc_1$) & CJG & RNB & LO \\
\hline
\multirow{5}{*}{\makecell[c]{ResNet-18\\(CIFAR-10)\\$Acc_O$: 93.14\%}} & Full GC-Net & 10.0 & 10.0 & 10.0 & 10.0 \\
& GC-Net-FH & 60.96 & 40.46 & 15.84 & 11.61 \\
& GC-Net-BH & 48.49 & 38.14 & 13.11 & 13.52 \\
& GC-Net-B25\% & 86.46 & 67.0 & 21.24 & 14.12 \\
& Original & \cellcolor{yellow!25}\textbf{90.55} & \cellcolor{yellow!25}\textbf{70.34} & \cellcolor{yellow!25}\textbf{24.92} & \cellcolor{yellow!25}\textbf{14.31} \\
\hline
\multirow{5}{*}{\makecell[c]{ResNet-18\\(FMNIST)\\$Acc_O$: 93.63\%}} & Full GC-Net & 90.59 & 21.67 & 46.88 & 22.15 \\
& GC-Net-FH & 90.55 & 24.95 & 45.48 & 24.85 \\
& GC-Net-BH & 91.96 & 30.87 & 65.19 & \textbf{28.79} \\
& GC-Net-B25\% & 92.14 & 33.96 & \textbf{65.27} & 24.78 \\
& Original & \cellcolor{yellow!25}\textbf{92.38} & \cellcolor{yellow!25}\textbf{34.33} & \cellcolor{yellow!25}64.15 & \cellcolor{yellow!25}25.25 \\
\hline
\multirow{5}{*}{\makecell[c]{ResNet-18\\(Tiny-IN)\\$Acc_O$: 65.15\%}} & Full GC-Net & 23.86 & 2.01 & 1.45 & 1.88 \\
& GC-Net-FH & 48.69 & 5.07 & 2.53 & 4.03 \\
& GC-Net-BH & 35.31 & 4.47 & 2.08 & 3.01 \\
& GC-Net-B25\% & 46.34 & 6.2 & 2.24 & 5.21 \\
& Original & \cellcolor{yellow!25}\textbf{58.17} & \cellcolor{yellow!25}\textbf{7.64} & \cellcolor{yellow!25}\textbf{3.11} & \cellcolor{yellow!25}\textbf{5.97} \\
\hline
\end{tabular}
\caption{ResNet-18 \textbf{OS-SynFlow} average accuracy (\%) across datasets and hybrid types at \textbf{80\% sparsity} using cosine similarity. Results for applying the pruning method directly to the original model, without a hybrid method, are highlighted in yellow.}
\label{tab:performance_comparison_resnet18_OS_SynFlow_cosine_80}
\end{table}

\begin{table}[!ht]
\centering
\setlength{\tabcolsep}{4pt}
\footnotesize
\begin{tabular}{|>{\centering\arraybackslash}p{1.8cm}|c|cccc|}
\hline
\multirow{2}{*}{\makecell[c]{Architecture}} & \multirow{2}{*}{Hybrid Type} & \multicolumn{4}{c|}{\textbf{C-SNIP - Sparsity 20\%}} \\
 &  & Orig($Acc_1$) & CJG & RNB & LO \\
\hline
\multirow{5}{*}{\makecell[c]{VGG16-BN\\(CIFAR-10)\\ $Acc_O$: 93.18\%}} & Full GC-Net & 92.77 & 74.88 & \textbf{29.97} & 13.11 \\
& GC-Net-FH & 93.08 & 76.03 & 29.45 & 13.02 \\
& GC-Net-BH & 92.95 & \textbf{76.55} & 29.83 & 12.91 \\
& GC-Net-B25\% & \textbf{93.29} & 76.29 & 29.77 & \textbf{13.72} \\
& Original & \cellcolor{yellow!25}93.23 & \cellcolor{yellow!25}76.41 & \cellcolor{yellow!25}29.62 & \cellcolor{yellow!25}13.18 \\
\hline
\multirow{5}{*}{\makecell[c]{VGG16-BN\\(FMNIST)\\ $Acc_O$: 94.36\%}} & Full GC-Net & 94.06 & 32.11 & 63.85 & 23.41 \\
& GC-Net-FH & 94.18 & \textbf{36.05} & \textbf{66.84} & 24.84 \\
& GC-Net-BH & 94.04 & 33.45 & 65.59 & 24.64 \\
& GC-Net-B25\% & 94.13 & 35.15 & 65.94 & 26.03 \\
& Original & \cellcolor{yellow!25}\textbf{94.22} & \cellcolor{yellow!25}34.03 & \cellcolor{yellow!25}65.31 & \cellcolor{yellow!25}\textbf{26.23} \\
\hline
\multirow{5}{*}{\makecell[c]{VGG16-BN\\(Tiny-IN)\\ $Acc_O$: 64.40\%}} & Full GC-Net & 60.68 & 6.08 & 4.82 & 4.52 \\
& GC-Net-FH & 58.5 & 6.05 & 4.19 & 4.4 \\
& GC-Net-BH & 61.62 & 6.52 & 5.29 & 4.66 \\
& GC-Net-B25\% & \textbf{63.25} & \textbf{6.71} & \textbf{5.56} & 4.67 \\
& Original & \cellcolor{yellow!25}63.03 & \cellcolor{yellow!25}6.5 & \cellcolor{yellow!25}5.35 & \cellcolor{yellow!25}\textbf{4.89} \\
\hline
\end{tabular}
\caption{VGG16-BN \textbf{C-SNIP} average accuracy (\%) across datasets and hybrid types at \textbf{20\% sparsity} using cosine similarity. Results for applying the pruning method directly to the original model, without a hybrid method, are highlighted in yellow.}
\label{tab:performance_comparison_vgg16bn_C-SNIP_20_cosine}
\end{table}

\begin{table}[!ht]
\centering
\setlength{\tabcolsep}{4pt}
\footnotesize
\begin{tabular}{|>{\centering\arraybackslash}p{1.8cm}|c|cccc|}
\hline
\multirow{2}{*}{\makecell[c]{Architecture}} & \multirow{2}{*}{Hybrid Type} & \multicolumn{4}{c|}{\textbf{C-SNIP - Sparsity 40\%}} \\
 &  & Orig($Acc_1$) & CJG & RNB & LO \\
\hline
\multirow{5}{*}{\makecell[c]{VGG16-BN\\(CIFAR-10)\\ $Acc_O$: 93.18\%}} & Full GC-Net & 92.5 & 75.52 & 29.53 & 13.02 \\
& GC-Net-FH & 92.7 & 75.51 & \textbf{29.86} & 13.01 \\
& GC-Net-BH & 92.87 & 75.76 & 29.56 & 13.12 \\
& GC-Net-B25\% & 93.19 & \textbf{76.84} & 29.78 & 13.4 \\
& Original & \cellcolor{yellow!25}\textbf{93.2} & \cellcolor{yellow!25}76.21 & \cellcolor{yellow!25}29.47 & \cellcolor{yellow!25}\textbf{13.45} \\
\hline
\multirow{5}{*}{\makecell[c]{VGG16-BN\\(FMNIST)\\ $Acc_O$: 94.36\%}} & Full GC-Net & 93.86 & 32.03 & 62.89 & 22.05 \\
& GC-Net-FH & 94.17 & 32.91 & 65.03 & 17.2 \\
& GC-Net-BH & 94.08 & 33.57 & \textbf{65.52} & 25.23 \\
& GC-Net-B25\% & 94.09 & \textbf{34.91} & 65.43 & 24.67 \\
& Original & \cellcolor{yellow!25}\textbf{94.24} & \cellcolor{yellow!25}34.02 & \cellcolor{yellow!25}64.81 & \cellcolor{yellow!25}\textbf{26.51} \\
\hline
\multirow{5}{*}{\makecell[c]{VGG16-BN\\(Tiny-IN)\\ $Acc_O$: 64.40\%}} & Full GC-Net & 59.17 & 6.02 & 4.79 & 4.49 \\
& GC-Net-FH & 53.78 & 5.56 & 4.41 & 4.15 \\
& GC-Net-BH & 60.38 & \textbf{6.79} & 5.09 & \textbf{5.05} \\
& GC-Net-B25\% & 62.81 & 6.6 & \textbf{5.47} & 4.98 \\
& Original & \cellcolor{yellow!25}\textbf{62.89} & \cellcolor{yellow!25}6.46 & \cellcolor{yellow!25}5.4 & \cellcolor{yellow!25}4.68 \\
\hline
\end{tabular}
\caption{VGG16-BN \textbf{C-SNIP} average accuracy (\%) across datasets and hybrid types at \textbf{40\% sparsity} using cosine similarity. Results for applying the pruning method directly to the original model, without a hybrid method, are highlighted in yellow.}
\label{tab:performance_comparison_vgg16bn_C-SNIP_40_cosine}
\end{table}

\begin{table}[!ht]
\centering
\setlength{\tabcolsep}{4pt}
\footnotesize
\begin{tabular}{|>{\centering\arraybackslash}p{1.8cm}|c|cccc|}
\hline
\multirow{2}{*}{\makecell[c]{Architecture}} & \multirow{2}{*}{Hybrid Type} & \multicolumn{4}{c|}{\textbf{C-SNIP - Sparsity 60\%}} \\
 &  & Orig($Acc_1$) & CJG & RNB & LO \\
\hline
\multirow{5}{*}{\makecell[c]{VGG16-BN\\(CIFAR-10)\\ $Acc_O$: 93.18\%}} & Full GC-Net & 91.73 & 75.17 & 29.13 & 12.58 \\
& GC-Net-FH & 91.99 & 75.29 & \textbf{29.29} & 12.68 \\
& GC-Net-BH & 92.19 & 75.36 & 29.25 & 12.68 \\
& GC-Net-B25\% & 92.47 & \textbf{76.27} & 29.24 & 12.89 \\
& Original & \cellcolor{yellow!25}\textbf{92.58} & \cellcolor{yellow!25}75.81 & \cellcolor{yellow!25}29.12 & \cellcolor{yellow!25}\textbf{13.06} \\
\hline
\multirow{5}{*}{\makecell[c]{VGG16-BN\\(FMNIST)\\ $Acc_O$: 94.36\%}} & Full GC-Net & 93.74 & 32.91 & 61.97 & 21.02 \\
& GC-Net-FH & 93.89 & 33.55 & 63.31 & 23.61 \\
& GC-Net-BH & 93.96 & 33.93 & 64.23 & 22.14 \\
& GC-Net-B25\% & 94.02 & \textbf{34.48} & \textbf{64.56} & 24.27 \\
& Original & \cellcolor{yellow!25}\textbf{94.09} & \cellcolor{yellow!25}34.37 & \cellcolor{yellow!25}63.72 & \cellcolor{yellow!25}\textbf{25.71} \\
\hline
\multirow{5}{*}{\makecell[c]{VGG16-BN\\(Tiny-IN)\\ $Acc_O$: 64.40\%}} & Full GC-Net & 58.24 & 5.85 & 4.69 & 4.32 \\
& GC-Net-FH & 52.09 & 5.39 & 4.28 & 4.07 \\
& GC-Net-BH & 58.77 & 6.29 & 5.12 & 4.43 \\
& GC-Net-B25\% & 60.41 & \textbf{6.48} & \textbf{5.31} & \textbf{5.03} \\
& Original & \cellcolor{yellow!25}\textbf{60.82} & \cellcolor{yellow!25}6.29 & \cellcolor{yellow!25}5.13 & \cellcolor{yellow!25}4.74 \\
\hline
\end{tabular}
\caption{VGG16-BN \textbf{C-SNIP} average accuracy (\%) across datasets and hybrid types at \textbf{60\% sparsity} using cosine similarity. Results for applying the pruning method directly to the original model, without a hybrid method, are highlighted in yellow.}
\label{tab:performance_comparison_vgg16bn_C-SNIP_60_cosine}
\end{table}

\begin{table}[!ht]
\centering
\setlength{\tabcolsep}{4pt}
\footnotesize
\begin{tabular}{|>{\centering\arraybackslash}p{1.8cm}|c|cccc|}
\hline
\multirow{2}{*}{\makecell[c]{Architecture}} & \multirow{2}{*}{Hybrid Type} & \multicolumn{4}{c|}{\textbf{C-SNIP - Sparsity 80\%}} \\
 &  & Orig($Acc_1$) & CJG & RNB & LO \\
\hline
\multirow{5}{*}{\makecell[c]{VGG16-BN\\(CIFAR-10)\\ $Acc_O$: 93.18\%}} & Full GC-Net & 90.34 & 73.52 & 28.85 & 12.16 \\
& GC-Net-FH & 90.61 & 73.99 & \textbf{28.92} & 12.28 \\
& GC-Net-BH & 90.81 & 74.08 & 28.84 & 12.39 \\
& GC-Net-B25\% & 91.07 & \textbf{74.39} & 28.73 & 12.49 \\
& Original & \cellcolor{yellow!25}\textbf{91.11} & \cellcolor{yellow!25}74.25 & \cellcolor{yellow!25}28.65 & \cellcolor{yellow!25}\textbf{12.63} \\
\hline
\multirow{5}{*}{\makecell[c]{VGG16-BN\\(FMNIST)\\ $Acc_O$: 94.36\%}} & Full GC-Net & 93.44 & 32.22 & 60.96 & 20.51 \\
& GC-Net-FH & 93.65 & 32.63 & 63.16 & 21.98 \\
& GC-Net-BH & 93.75 & 32.96 & 63.93 & 21.5 \\
& GC-Net-B25\% & 93.86 & 33.19 & \textbf{64.25} & 22.38 \\
& Original & \cellcolor{yellow!25}\textbf{93.89} & \cellcolor{yellow!25}\textbf{33.45} & \cellcolor{yellow!25}63.83 & \cellcolor{yellow!25}\textbf{22.88} \\
\hline
\multirow{5}{*}{\makecell[c]{VGG16-BN\\(Tiny-IN)\\ $Acc_O$: 64.40\%}} & Full GC-Net & 57.43 & 5.67 & 4.55 & 4.2 \\
& GC-Net-FH & 50.62 & 5.21 & 4.16 & 4.06 \\
& GC-Net-BH & 57.51 & 6.03 & 4.93 & 4.28 \\
& GC-Net-B25\% & 58.91 & \textbf{6.22} & \textbf{5.09} & 4.43 \\
& Original & \cellcolor{yellow!25}\textbf{59.05} & \cellcolor{yellow!25}6.11 & \cellcolor{yellow!25}5.01 & \cellcolor{yellow!25}\textbf{4.52} \\
\hline
\end{tabular}
\caption{VGG16-BN \textbf{C-SNIP} average accuracy (\%) across datasets and hybrid types at \textbf{80\% sparsity} using cosine similarity. Results for applying the pruning method directly to the original model, without a hybrid method, are highlighted in yellow.}
\label{tab:performance_comparison_vgg16bn_C-SNIP_80_cosine}
\end{table}

\clearpage

\subsection{GC-Net Creation Algorithm}
\label{subsec9}
\begin{algorithm}[h!]
    \footnotesize
    \SetAlgoLined
\caption{GC-Net Creation}
\label{alg:GC-Net Creation}
Given $F_O$, load weights $W_O$
\begin{center}
\vspace{-0.5cm}
\begin{tcolorbox}[colback=blue!4!white,colframe=blue!75!black]
\vspace{-.1in}
Activation Matrix Calculation:\\
\For{each sample $s \in S$}
{Pass $s$ through the trained network\\
\For{each layer $l \in \{1, \ldots, L\}$}
{Collect activation state $A_l[s, o, h, w]$}
}
\For{each layer $l \in \{1, \ldots, L\}$ in $F_O$}
{Initialize activation matrix $M_l[s, o] = 0$\\
\For{each sample $s \in S$}
{\For{each output channel $o$}
{$M_l[s, o] = \frac{1}{h \cdot w} \sum_{i=1}^h \sum_{j=1}^w A_l[s, o, i, j]$}
}
}
\vspace{-.1in}
\end{tcolorbox}
\end{center}
\vspace{-0.4cm}
\begin{center}
\vspace{-0.5cm}
\begin{tcolorbox}[colback=blue!8!white,colframe=blue!75!black]
\vspace{-.1in}
Connectivity Matrix Calculation:\\
\For{$l = 1$ \KwTo $L$}
{$R_l \leftarrow \text{zeros}(o_{l+1}, o_l)$\\
\For{$i = 1$ \KwTo $o_{l+1}$}
{\For{$j = 1$ \KwTo $o_l$}
{$R_l[i,j] \leftarrow \rho(M_l[:,j], M_{l+1}[:,i])$}}}
\vspace{-.1in}
\end{tcolorbox}
\vspace{-0.4cm}
\end{center}
\begin{center}
\vspace{-0.3cm}
\begin{tcolorbox}[colback=blue!12!white,colframe=blue!65!black]
\vspace{-.1in}
Connectivity Matrix Expansion:\\
\For{$l = 1$ \KwTo $L$}
{$[a, b, c, c] \leftarrow \text{Dimensions}(\text{Layer}_{l+1})$\\
$[a, g, c, c] \leftarrow \text{Dimensions}(\text{Layer}_l)$\\
$R_l \leftarrow R_l^T$\\
$R_l \leftarrow \text{Reshape}(R_l, [a, g, c, c])$}
\vspace{-.1in}
\end{tcolorbox}
\vspace{-0.5cm}
\end{center}
Given $F_{GC}$, set layer $l = 1$ to an Identity layer\\
\For{$l = 2$ \KwTo $L$}
{Load $R_l$ as weights for $l+1$}
Save $F_{GC}$ weights, $W_{GC}$
\end{algorithm}


{\small
\bibliographystyle{ieee_fullname}
\bibliography{GC-Net-arXiv}
}

\end{document}